%% file: main.tex
\pdfoutput=1
\RequirePackage{fix-cm}
\documentclass[twocolumn]{svjour3}          % twocolumn
\smartqed  % flush right qed marks, e.g. at end of proof
\usepackage{graphicx}
\journalname{ } % The correct name will be entered by the editor

\input{macros}

\begin{document}

\title{FPO++: Efficient Encoding and Rendering of Dynamic Neural Radiance Fields by Analyzing and Enhancing Fourier PlenOctrees}
\titlerunning{FPO++}
\subtitle{}
\author{Saskia Rabich \and Patrick Stotko \and Reinhard Klein}
\institute{Saskia Rabich (Corr. Author: srabich@cs.uni-bonn.de) \at University of Bonn, Germany \and Patrick Stotko \at University of Bonn, Germany \and Reinhard Klein \at University of Bonn, Germany}
\date{ }% The correct dates will be entered by the editor

\maketitle

\begin{abstract}
    Fourier PlenOctrees have shown to be an efficient representation for real-time rendering of dynamic Neural Radiance Fields (NeRF).
    Despite its many advantages, this method suffers from artifacts introduced by the involved compression when combining it with recent state-of-the-art techniques for training the static per-frame NeRF models.
    In this paper, we perform an in-depth analysis of these artifacts and leverage the resulting insights to propose an improved representation.
    In particular, we present a novel density encoding that adapts the Fourier-based compression to the characteristics of the transfer function used by the underlying volume rendering procedure and leads to a substantial reduction of artifacts in the dynamic model.
    We demonstrate the effectiveness of our enhanced Fourier PlenOctrees in the scope of quantitative and qualitative evaluations on synthetic and real-world scenes.

\keywords{Neural Radiance Fields \and Dynamic Scenes \and Real-time Rendering \and Encoding \and Fourier Transform}
\end{abstract}

\input{sections/introduction}

\input{sections/relatedwork}

\input{sections/preliminaries}

\input{sections/method}

\input{sections/results}

\input{sections/conclusion}

\section*{Acknowledgements}
This work has been funded by the Federal Ministry of Education and Research under grant no. 01IS22094E WEST-AI, by the Federal Ministry of Education and Research of Germany and the state of North-Rhine Westphalia as part of the Lamarr-Institute for Machine Learning and Artificial Intelligence, and additionally by the DFG project KL 1142/11-2 (DFG Research Unit FOR 2535 Anticipating Human Behavior).

% Note: If you are using BibTeX, please use the following code:
\bibliographystyle{spmpsci}
\bibliography{refs} 
% where "bib_CGIconf" has to be replaced by the name of your BibTeX
% file (without the .bib extension). 

% Non-BibTeX users please use:
%\begin{thebibliography}{}
% and use \bibitem to create references as below. For the names of
% the authors, please write each name followed by the initials,
% e.g.: Knuth D.E., Lamport L.
%\bibitem{RefA}
%\end{thebibliography}

\end{document}

%% file: macros.tex
\usepackage{amsmath}
\usepackage{amssymb}
\usepackage{caption}
\usepackage{subcaption}
\usepackage{booktabs}
\usepackage{rotating}
\usepackage{float}
\usepackage{tikz}
\usepackage{url}

\usetikzlibrary {arrows.meta} 
\renewcommand{\vec}[1]{\boldsymbol{#1}}

\renewcommand{\vec}[1]{\mathbf{#1}}

\newcommand{\norm}[1]{\left\Vert#1\right\Vert}

\let\originalleft\left
\let\originalright\right
\renewcommand{\left}{\mathopen{}\mathclose\bgroup\originalleft}
\renewcommand{\right}{\aftergroup\egroup\originalright}

\definecolor{best}{rgb}{0.8,1.0,0.7}
\definecolor{second}{rgb}{1.0,0.97,0.7}

%% file: sections/introduction.tex
\begin{tikzpicture}[remember picture,overlay]%
\node[anchor=south,align=center,font=\mdseries,yshift=0.8cm, xshift=-0.3cm] at (current page.south) {%
\begin{minipage}[c]{\textwidth}This version of the article has been accepted for publication, after peer review (when applicable) but is not the Version of Record and does not reflect post-acceptance improvements, or any corrections. The Version of Record is available online at: \url{http://dx.doi.org/10.1007/s00371-024-03475-3}\end{minipage}%
};%
\end{tikzpicture}%

\section{Introduction}

Photorealistic rendering of dynamic real-world scenes such as moving persons or interactions of people with surrounding objects plays a vital role in 4D content generation and has numerous applications including augmented reality (AR) and virtual reality (VR), advertisement, or entertainment.
Traditional approaches typically capture such scenarios with professional well-calibrated hardware setups~\cite{collet2015high,guo2019relightables} in a controlled environment.
This way, high-fidelity reconstructions of scene geometry, material properties, and surrounding illumination can be obtained.
Recent advances in neural scene representations and, in particular, the seminal work in Neural Radiance Fields (NeRF)~\cite{mildenhall2020nerf} marked a breakthrough towards synthesizing photorealistic novel views.
Unlike in previous approaches, highly detailed renderings of complex static scenes can be generated only from a set of posed multi-view images recorded by commodity cameras.
Several extensions have subsequently been developed to alleviate the limitations of the original NeRF approach which led to significant reductions in the training times~\cite{mueller2022instant} or acceleration of the rendering performance~\cite{garbin2021fastnerf,chen2023mobilenerf}.

Further approaches explored the application of NeRF to dynamic scenarios but still suffer from slow rendering speed~\cite{guo2022neural,liu2022devrf,li2022neural,song2023nerfplayer}.
Among these, Fourier PlenOctrees (FPO)~\cite{wang2022fourier} offer an efficient representation and compression of the temporally evolving scene while at the same time enabling free viewpoint rendering in real time.
In particular, they join the advantages of the static PlenOctree representation~\cite{yu2021plenoctrees} with a Discrete Fourier Transform (DFT) compression technique to compactly store time-dependent information in a sparse octree structure.
Although this elegant formulation enables a high runtime performance, the Fourier-based compression results in a low-frequency approximation of the original data.
This estimate is susceptible to artifacts in both the reconstructed geometry and color of the model which often persist and cannot be fully resolved even after an additional fine-tuning step.
Strong priors like a pretrained generalizable NeRF~\cite{wang2021ibrnet} may mitigate these artifacts and are applied in FPOs for a more robust initialization.
However, when considering recent state-of-the-art techniques to boost the training of the static per-frame neural radiance fields without requiring prior knowledge, obtaining a suitable compressed model remains challenging.

In this paper, we revisit the frequency-based compression of Fourier PlenOctrees in the context of volume rendering and investigate the characteristics and behavior of the involved time-dependent density functions.
Our analysis reveals that they exhibit beneficial properties after the decompression that can be exploited via the implicit clipping behavior in terms of an additional Rectified Linear Unit (ReLU) operation applied for rendering that enforces non-negative values.
Based on these observations, we aim to find efficient strategies that retain the compact representation of FPOs without introducing significant additional complexity or overhead while eliminating artifacts and even further accelerating the high rendering performance.
We particularly focus on flexible approaches that allow for interchanging components to leverage recent advances and, thus, model the Fourier-based compression as an explicit step in the training process, instead of investigating end-to-end trainable systems.
To this end, we derive an efficient density encoding consisting of two transformations, where 1) a component-dependent encoding counteracts the under-estimation of values inherent to the reconstruction with a reduced set of Fourier coefficients, and 2) a further logarithmic encoding facilitates the reconstruction from Fourier coefficients and the fine-tuning process by putting higher attention to small values in the underlying $ L_2 $-minimization.
While it is tempting to learn such an encoding in an end-to-end fashion using e.g. small MLPs, our handcrafted encoding directly incorporates the gained insights and only adds negligible computation overhead, which is especially beneficial for fast rendering.

In summary, our key contributions are:
\begin{itemize}
    \item We perform an in-depth analysis of the compression artifacts induced in the Fourier PlenOctree representation when recent state-of-the-art techniques for training the individual per-frame NeRF models are employed.
    \item We introduce a novel density encoding for the Fourier-based compression that adapts to the characteristics of the transfer function in volume rendering of NeRF.
\end{itemize}
%
%Our source code is available at \url{https://github.com/SaskiaRabich/FPOplusplus}.

%% file: sections/relatedwork.tex
\section{Related Work}

\subsection{Neural Scene Representations}

With the rise of machine learning, neural networks have become increasingly popular and attracted significant interest for reconstructing and representing scenes~\cite{sitzmann2019scene,mildenhall2019local,bi2020deep,niemeyer2020differentiable,yariv2020multiview,yariv2021volume,jena2022neural,suhail2022light}.
In this context, the seminal work on NeRF~\cite{mildenhall2020nerf} demonstrated impressive results for synthesizing novel views using volume rendering of density and view-dependent radiance that are optimized in an implicit neural scene representation using multi-layer perceptrons (MLP) from a set of posed images.
Further work focused on addressing several of its limitations such as relaxing the dependence on accurate camera poses by jointly optimizing extrinsic~\cite{wang2021nerf,lin2021barf,wang2023badnerf} and intrinsic parameters~\cite{wang2021nerf,wang2023badnerf}, the dependence on sharp input images by a simulation of the blurring process~\cite{ma2022deblur,wang2023badnerf}, representing unbounded scenes~\cite{zhang2020nerf++,barron2022mip} and large-scene environments~\cite{tancik2022block}, or reducing aliasing artifacts by modeling the volumetric rendering with conical frustums instead of rays~\cite{barron2021mip,barron2022mip,barron2023zipnerf}.

\subsection{Acceleration of NeRF Training and Rendering}

A further major limitation of implicit neural scene representations is the high computational cost during both training and rendering.
In order to speed up the rendering process, several techniques were proposed that reduce the amount of required samples along the ray~\cite{lindell2021autoint,neff2021donerf,kurz2022adanerf} or subdivide the scene and use smaller and faster networks for the evaluation of the individual parts~\cite{rebain2021derf,reiser2021kilonerf}.
Some approaches represented the scene using a latent feature embedding where feature vectors are stored in voxel grids~\cite{wu2022diver,sun2022direct} or octrees~\cite{liu2020neural}.
Another strategy for accelerating rendering relies on storing precomputed features efficiently into discrete representations such as sparse grids with a texture atlas~\cite{hedman2021baking}, textured polygon meshes~\cite{chen2023mobilenerf}, or caches~\cite{garbin2021fastnerf} and inferring view-dependent effects by a small MLP.
Furthermore, PlenOctrees~\cite{yu2021plenoctrees} use a hierarchical octree structure of the density and the view-dependent radiance in terms of spherical harmonics (SH) to entirely avoid network evaluations.

Improving the convergence of the training process has also been investigated by using additional data such as depth maps~\cite{deng2022depth} or a visual hull computed from binary foreground masks~\cite{kondo2021vaxnerf} as an additional guidance.
Furthermore, meta learning approaches allow for a more effective initialization compared to random weights~\cite{tancik2021learned}.
Similar to the advances in rendering performance, discrete scene representations were also leveraged to boost the training.
Instant-NGP~\cite{mueller2022instant} incorporated a multi-resolution hash encoding to significantly accelerate the training of neural models including NeRF.
Plenoxels~\cite{fridovich2022plenoxels} stored SH and opacity values within a sparse voxel grid and TensoRF~\cite{chen2022tensorf} factorized dense voxel grids into multiple low-rank components.
However, all the aforementioned methods and representations are limited to static scenes only and do not take dynamic scenarios like motion into account.

\subsection{Dynamic Scene Representations}

Although novel views of scenes containing motions can be directly synthesized from the individual per-frame static models, significant effort has been spent into more efficient representations for neural rendering such as subdividing the scene into static and dynamic parts~\cite{lin2021deep,wu2020multi}, using point clouds~\cite{wu2020multi}, mixtures of volumetric primitives~\cite{lombardi2021mixture}, deformable human models~\cite{peng2021neural}, or encoding the dynamics with encoder-decoder architectures~\cite{lombardi2019neural,meka2020deep}.
Due to the success and representation power of Neural Radiance Fields, these developments also inspired recent extensions of NeRF to dynamic scenes.
Some methods leveraged the additional temporal information to perform novel-view synthesis from a single video of a moving camera instead of large collections of multi-view images~\cite{pumarola2021d,tretschk2021non,li2021neural,du2021neural,park2021nerfies,gafni2021dynamic,xu2021hnerf,weng2022humannerf}.
Among these, the reconstruction of humans also gained increasing interest where morphable~\cite{gafni2021dynamic} and implicit generative models~\cite{xu2021hnerf}, pre-trained features~\cite{wang2021ibutter}, or deformation fields~\cite{park2021nerfies,weng2022humannerf} were employed to regularize the reconstruction.
Furthermore, T{\"o}RF~\cite{attal2021torf} used time-of-flight sensor measurements as an additional source of information and DyNeRF~\cite{li2022neural} learned time-dependent latent codes to constrain the radiance field.
Another way of handling scene dynamics is through the decomposition into separate networks where each handles a specific part such as static and dynamic content~\cite{gao2021dynamic}, rigid and non-rigid motion~\cite{weng2022humannerf}, new areas~\cite{song2023nerfplayer}, or even only a single dynamic entity~\cite{zhang2021editable}.
Similarly, some methods reconstruct the scene in a canonical volume and model motion via a separate temporal deformation field~\cite{liu2022devrf,guo2022neural,fang2022fast} or residual fields~\cite{li2022streaming,wang2023neural}.
Discrete grid-based representations~\cite{chen2022tensorf} applied for accelerating static scene training have also been extended to factorize the 4D spacetime~\cite{cao2023hexplane,shao2023tensor4d,isik2023humanrf,fridovich2023k}.
In this context, Fourier PlenOctrees (FPO)~\cite{wang2022fourier} relaxed the limitation of ordinary PlenOctrees~\cite{yu2021plenoctrees} to only capture static scenes in a hierarchical manner by combining it with the Fourier transform which enables handling time-variant density and SH-based radiance in an efficient way.

%% file: sections/preliminaries.tex
\section{Preliminaries}

In this section, we revisit the method of representing a dynamic scene using a Fourier PlenOctree~\cite{wang2022fourier}, which extends the model-free, static, explicit PlenOctree representation~\cite{yu2021plenoctrees} for real-time rendering of NeRFs.
Given a set of $T$ individual PlenOctrees each corresponding to a frame in a dynamic time sequence, the construction of an FPO consists of two parts:
1) a structure unification of the $T$ static models, and 2) the computation of the DFT-compressed octree leaf entries, which will be discussed in more detail in Sections~\ref{sec:unification} and~\ref{sec:compression}, respectively.

In order to render an image of a scene at time step $ { t \in \{0,\dots,T-1\} } $, the color $\hat{\vec{C}}$ of a pixel is accumulated along the ray $ { \vec{r}(\tau) = \vec{o} + \tau \cdot \vec{d} \in \mathbb{R}^3 } $ with origin $ { \vec{o} \in \mathbb{R}^3 } $ at the camera, viewing direction $ { \vec{d} \in \mathbb{R}^3 } $ as well as step length $ \tau \in \mathbb{R}_{\ge 0} $. 
The ray $\vec{r}$ is taken from the set of all rays $ \mathcal{R} $ cast from the input images.
The accumulation is performed analogously to PlenOctrees~\cite{yu2021plenoctrees}:
\begin{equation}\label{eq:acccolor}
    \hat{\vec{C}}(\vec{r}, t) = \sum_{i=1}^{N} T_i(t) \, \left(1 - \exp(-\sigma_i(t) \, \delta_i)\right) \, \vec{c}_i(t),
\end{equation}
where $N$ is the number of octree leaves hit by $\vec{r}$, $ { \delta_i = \tau_{i + 1} - \tau_i } $ the distance between voxel borders and
\begin{equation}\label{eq:transmittance}
    T_i(t) = \exp\left( -\sum_{j=1}^{i-1} \sigma_j(t) \, \delta_j \right)
\end{equation}
is the accumulated transmittance from the camera up to the leaf node $i$. 
Hereby, $\left(1 - \exp(-\sigma_i(t) \, \delta_i)\right)$ can be considered as the transfer function from densities to transmittance in this node.

The rendering procedure of an FPO is analogous to PlenOctrees~\cite{yu2021plenoctrees}, with the addition of passing the time step $t$ to the renderer.
The time-dependent density $ { \sigma_i(t) \in \mathbb{R} } $ is reconstructed using the inverse discrete Fourier transform (IDFT) for time step $t$ applied to the values stored in the FPO in leaf node $i$.
Similarly, the time- and view-dependent color $ { \vec{c}_i(t) \in \mathbb{R}^3 } $ is obtained by first applying the IDFT to the FPO entries of the respective SH-coefficients $ { \vec{z}_i(t) \in \mathbb{R}^{Z \times 3} } $ with $Z$ SH-coefficients per color channel, and then querying $\vec{z}_i(t)$ for the given viewing direction $\vec{d}$.
Finally, the sigmoid function is applied to $ \vec{c}_i(t)$ for normalization.
In the following, we omit the subscript $i$ for brevity as all computations are performed per leaf.
Since all operations are differentiable with respect to the octree leaves, the compressed representation can be directly fine-tuned based on the rendered images using the following image loss function~\cite{wang2022fourier}:
\begin{equation}\label{eq:optimization}
    \mathcal{L} = \sum_{t = 0}^{T - 1} \sum_{\vec{r} \in \mathcal{R}} \norm{\vec{C}(\vec{r}, t) - \hat{\vec{C}}(\vec{r}, t)}_2^2.
\end{equation}

\subsection{PlenOctree Structure Unification}\label{sec:unification}

To construct an FPO, time-dependent SH coefficients and densities from all PlenOctrees are merged into a single data structure.
The sparse octree structures are first unified to obtain the structure of the FPO.
The static PlenOctrees contains leaves with maximum resolution only where the scene is non-empty. 
Identifying these regions over all time steps and refining the structure of all PlenOctrees accordingly yields the sparse octree structure for the dynamic representation~\cite{wang2022fourier}.

\subsection{Time-variant Data Compression}\label{sec:compression}

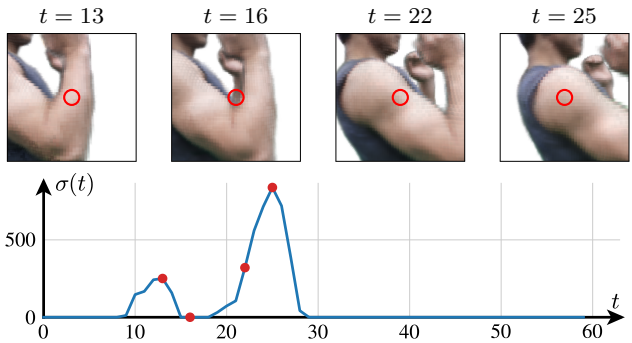
\begin{figure}
    {
    \setlength{\fboxsep}{0pt}
    \setlength{\fboxrule}{0.3pt}
    \centering
    \begin{minipage}{0.11\textwidth}
        \centering
        $t=13$\\
        \begin{tikzpicture}[
        node distance = 0mm,
        image/.style = {scale=1.,},
        ]
        \node[image] (A) {\fbox{\includegraphics[width=0.88\textwidth]{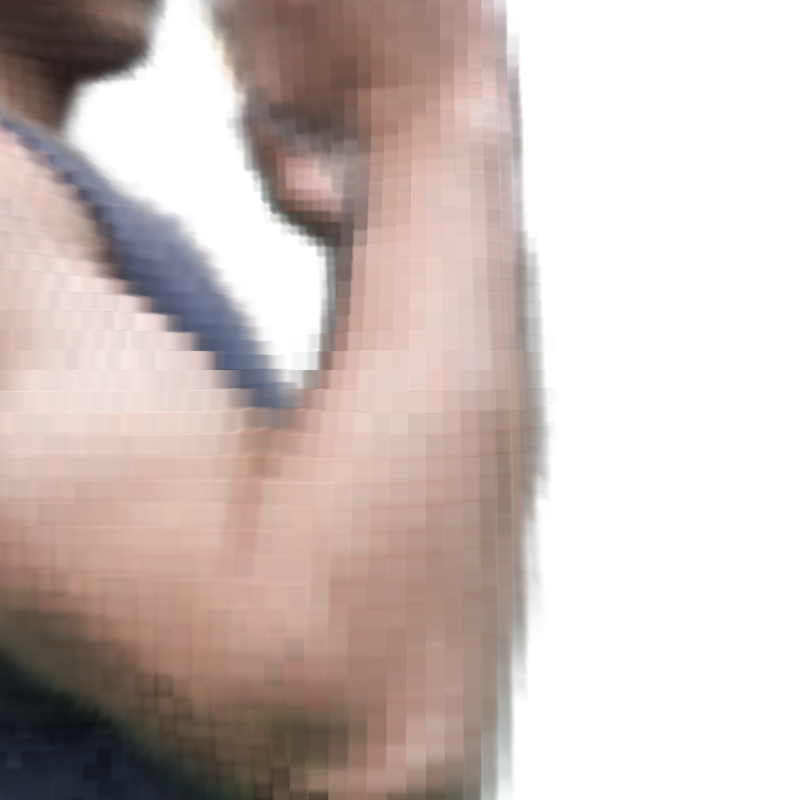}}};
        \draw(A.center)node[red,circle,draw,thick,inner sep=2pt](){};
        \end{tikzpicture}
    \end{minipage}
    \hfill
    \begin{minipage}{0.11\textwidth}
        \centering
        $t=16$\\
        \begin{tikzpicture}[
        node distance = 0mm,
        image/.style = {scale=1.,},
        ]
        \node[image] (A) {\fbox{\includegraphics[width=0.88\textwidth]{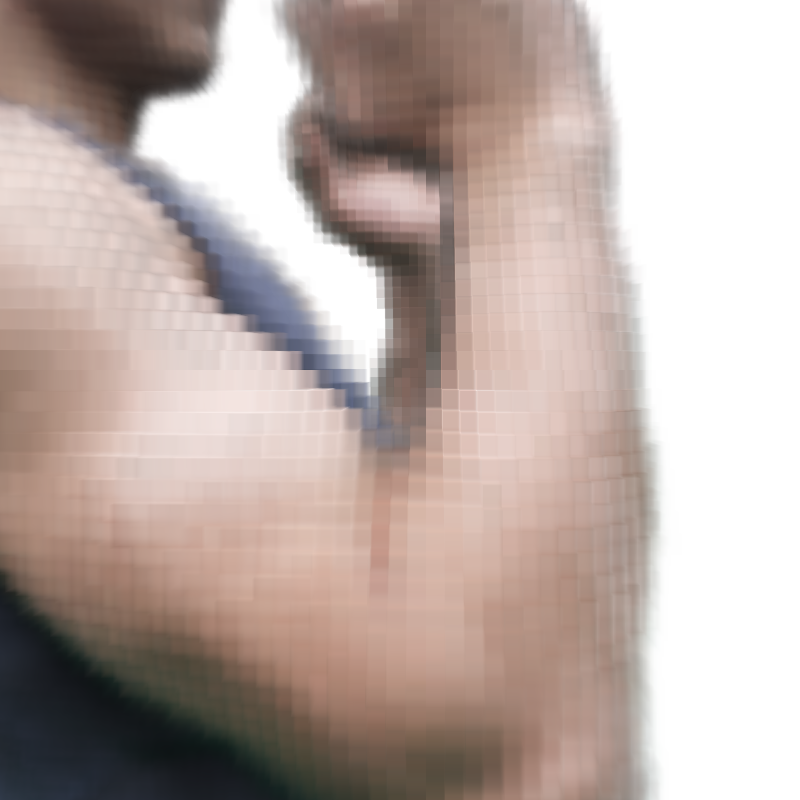}}};
        \draw(A.center)node[red,circle,draw,thick,inner sep=2pt](){};
        \end{tikzpicture}
    \end{minipage}
    \hfill
    \begin{minipage}{0.11\textwidth}
        \centering
        $t=22$\\
        \begin{tikzpicture}[
        node distance = 0mm,
        image/.style = {scale=1.,},
        ]
        \node[image] (A) {\fbox{\includegraphics[width=0.88\textwidth]{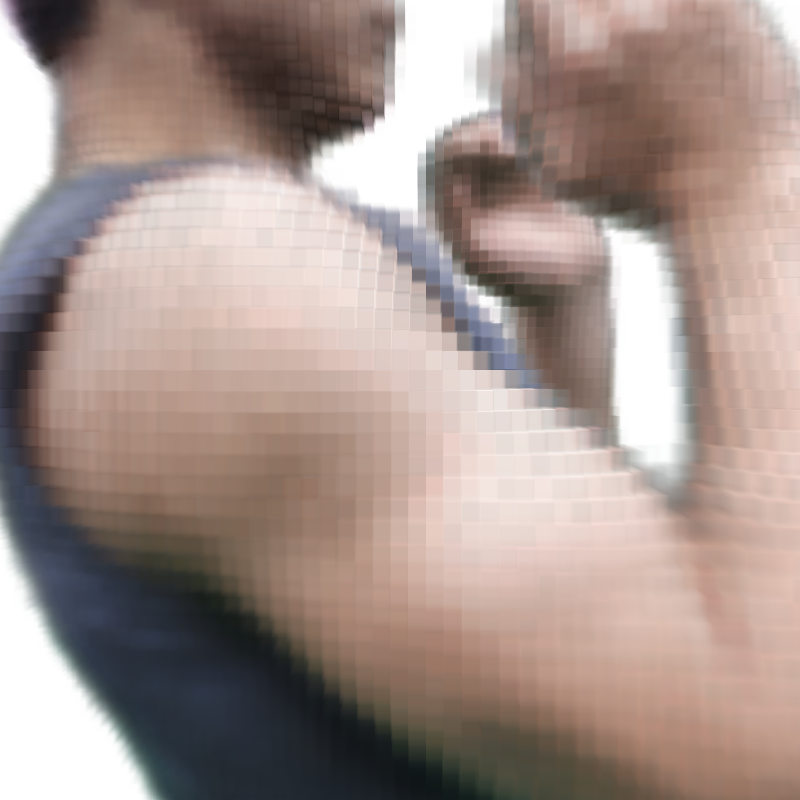}}};
        \draw(A.center)node[red,circle,draw,thick,inner sep=2pt](){};
        \end{tikzpicture}
    \end{minipage}
    \hfill
    \begin{minipage}{0.11\textwidth}
        \centering
        $t=25$\\
        \begin{tikzpicture}[
        node distance = 0mm,
        image/.style = {scale=1.,},
        ]
        \node[image] (A) {\fbox{\includegraphics[width=0.88\textwidth]{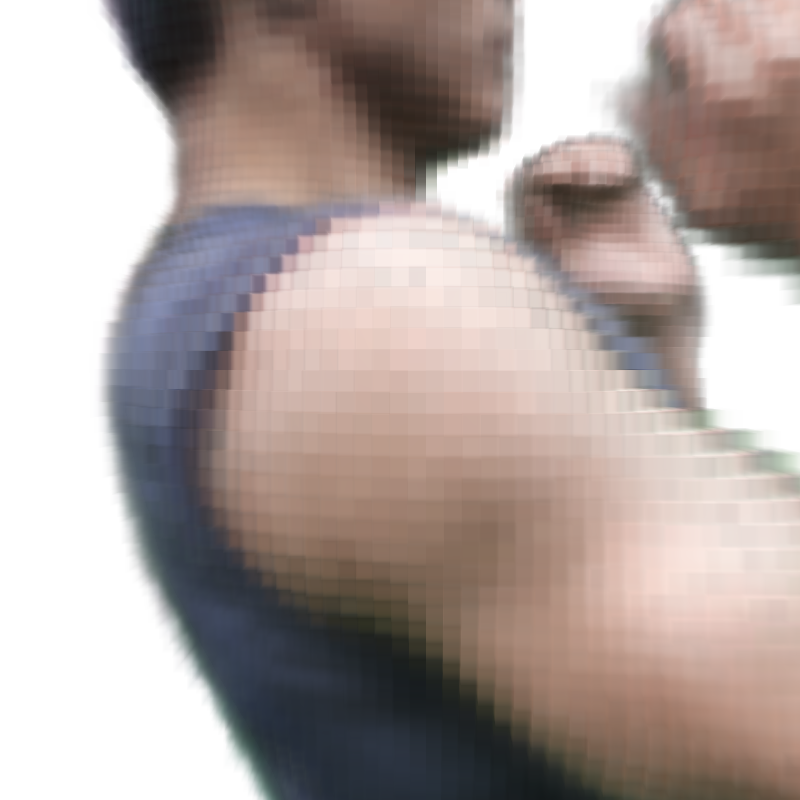}}};
        \draw(A.center)node[red,circle,draw,thick,inner sep=2pt](){};
        \end{tikzpicture}
    \end{minipage}\\
    \begin{minipage}{0.99\linewidth}
        \includegraphics[width=\textwidth]{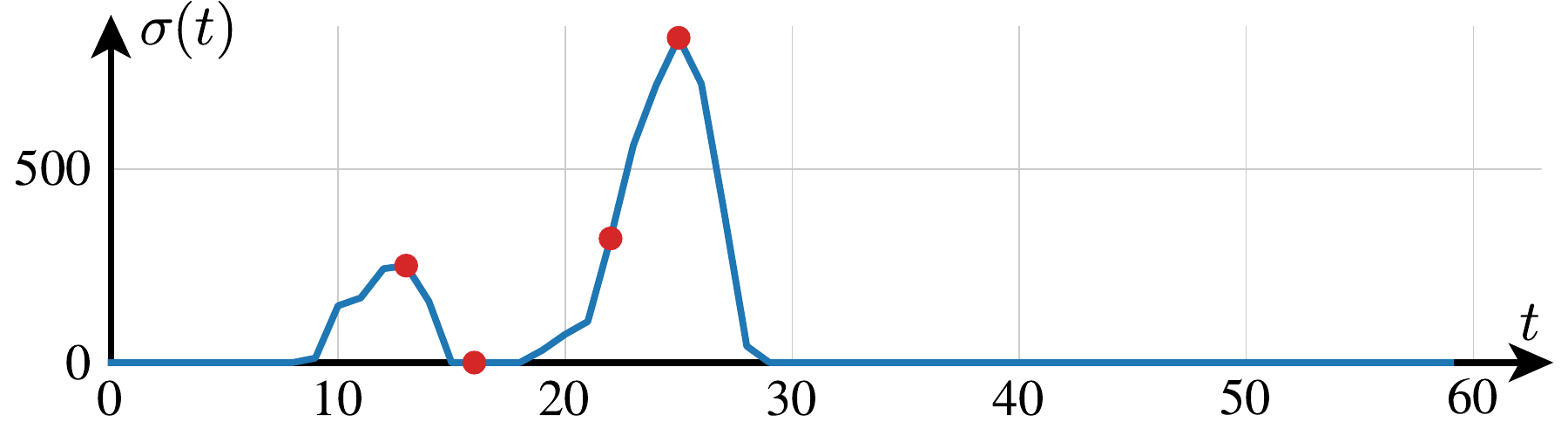}
    \end{minipage}
    }
    \caption{Density over time of a single octree leaf. The leaf is marked in red in the respective images taken from the same view at different time steps $t$. Although the opacity is similar in the views, highly varying densities are observed over time, except for $t=16$ where there is empty space in the tree leaf.}
    \label{fig:opacity}
\end{figure}

\begin{figure*}
    \centering
    \begin{minipage}{0.62\textwidth}
        \raggedleft
        \includegraphics[width=\textwidth]{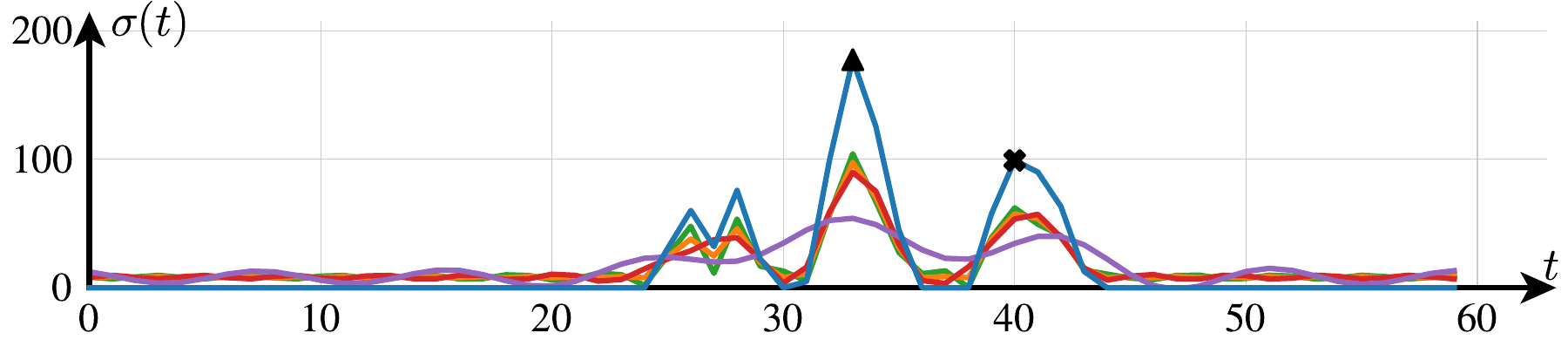}
        \includegraphics[width=0.99\textwidth]{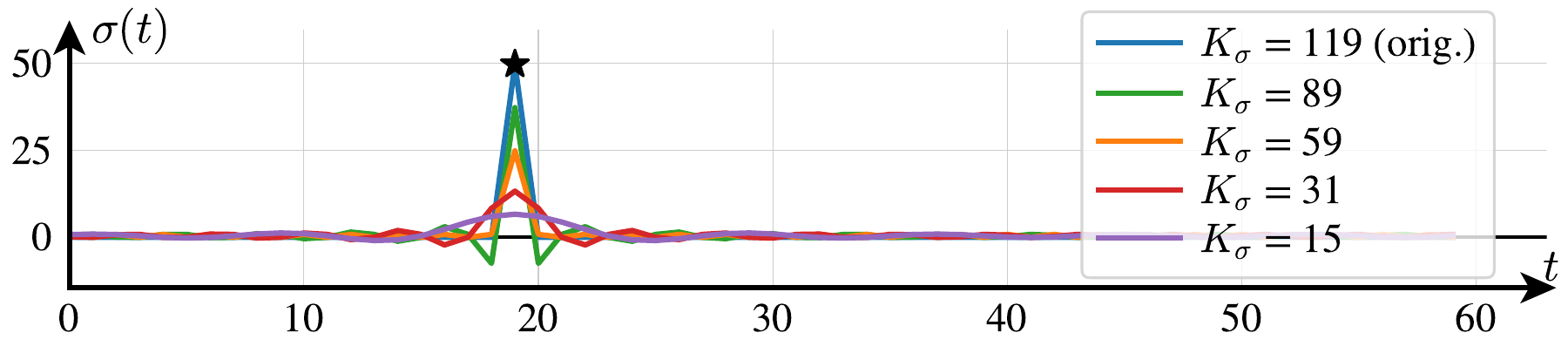}
    \end{minipage}\hfill
    \begin{minipage}{0.35\textwidth}
        \centering
        \includegraphics[width=\textwidth]{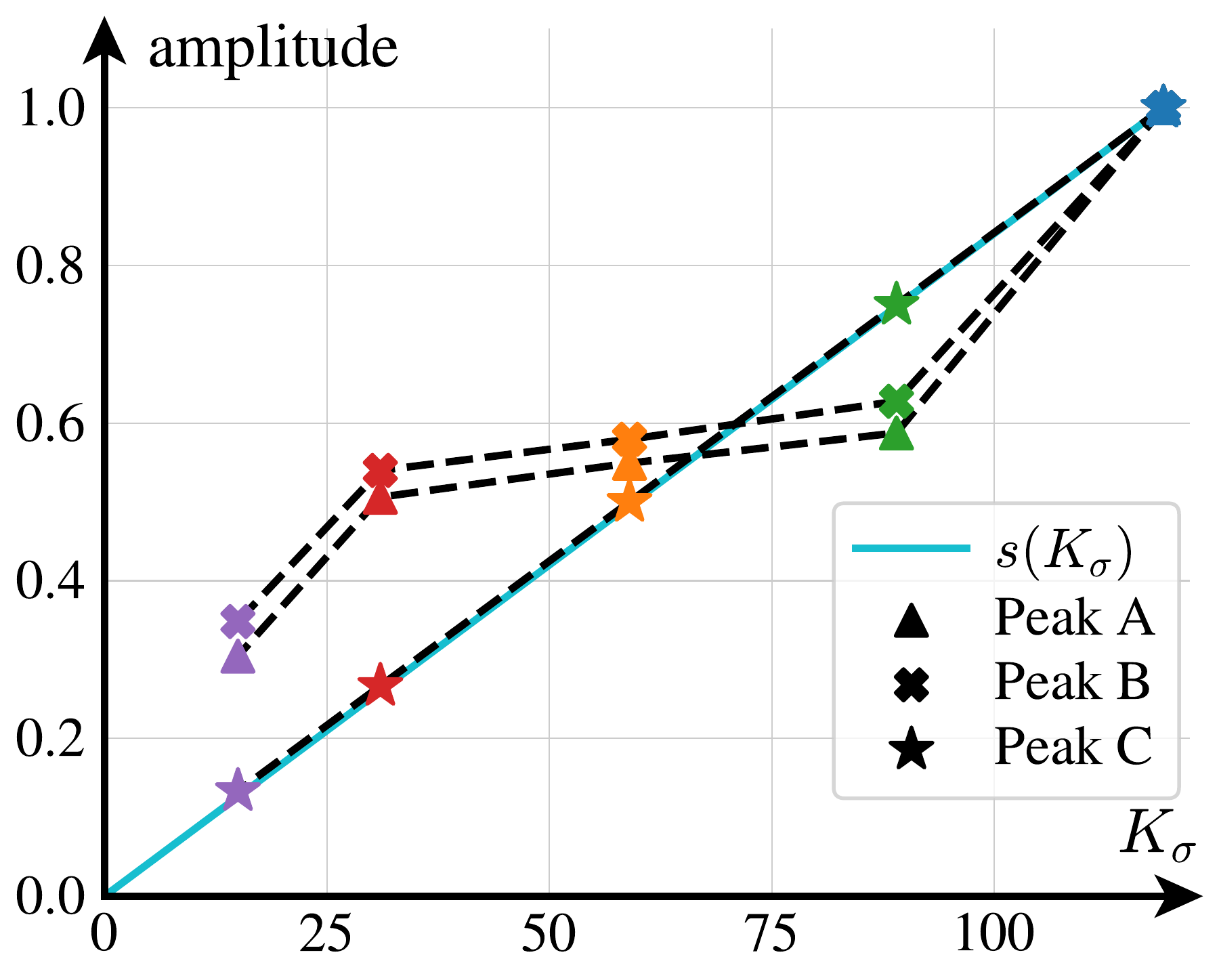}
    \end{minipage}
    \caption{Two exemplary density functions over time (left) reconstructed with different number of coefficients $K_\sigma$. The original function of $T$ time steps is equal to its reconstruction with $K_\sigma = 119$ Fourier coefficients. The falloff of the marked peaks relative to its original value (right) is depending on $K_\sigma$ and follows the linear scaling function $ s(K_\sigma) $.}
    \label{fig:comp_observation}
\end{figure*}

\begin{figure}
    \centering
    \begin{minipage}{0.47\textwidth}
    \centering
        \includegraphics[width=\textwidth]{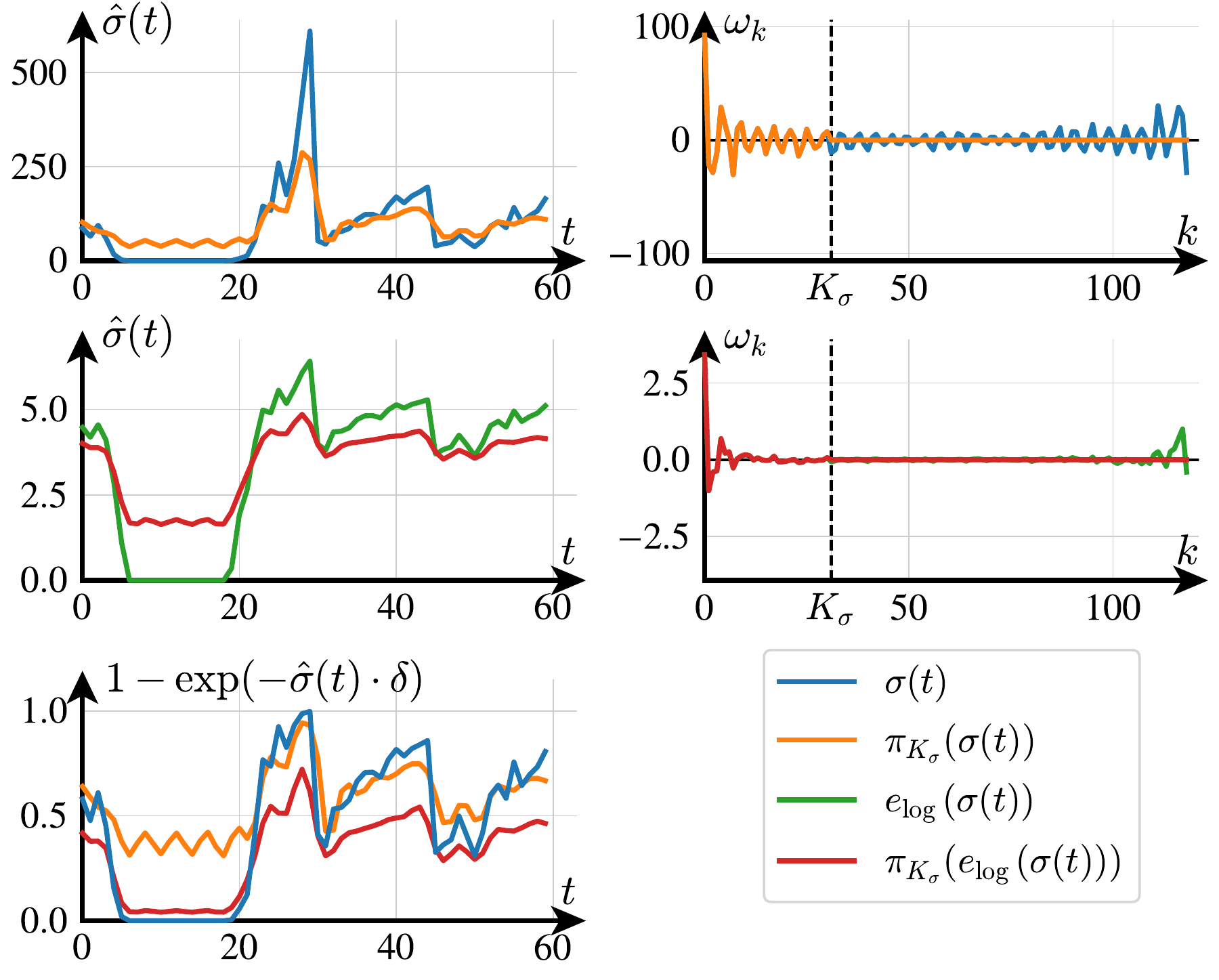}
    \end{minipage}
    \caption{A density function and its reconstruction $\pi_{K_\sigma}$ using the DFT and IDFT with ${K_\sigma=31}$ (top left) and the same function and its reconstruction after applying our logarithmic encoding $e_{\log}$ (center left). Their full and compressed Fourier representations (top right, center right) show that a logarithmically scaled function contains less high-frequency information that gets lost during compression. Applying the transfer function to the reconstructions (bottom left) shows that the logarithmic version can better represent the original one.} 
    \label{fig:saturation_observation}
\end{figure}

After creating the structure of the FPO, the SH coefficients and densities of all leaves and time steps are compressed by converting them into the frequency domain using the DFT.
Each SH coefficient and density value is compressed independently for each octree leaf, where only $ K_{\sigma} $ components for the transformed density functions and $ K_{\vec{z}} $ components for the SH coefficients are kept and stored.
Thereby, $ K $ components correspond to $ 0.5 \cdot (K + 1) $ frequencies and omitted components correspond to higher frequencies in the frequency domain, so a low-frequency approximation of the data is computed.
Thus, the entries of the Fourier PlenOctree are calculated as 
\begin{equation}\label{eq:DFT}
    \omega_k = \sum_{t=0}^{T-1} x(t) \cdot \textrm{DFT}_k(t)
\end{equation}
with
\begin{equation}
    \textrm{DFT}_k(t) = \begin{cases}
        \frac{1}{T} \cos\left(\frac{k \pi}{T} t\right) & \textrm{if $k$ is even,}\\
        \frac{1}{T} \sin\left(\frac{(k + 1) \pi}{T} t\right) & \textrm{if $k$ is odd.}
    \end{cases}
\end{equation}
Here, $ x $ represents either the density $ \sigma $ or a component of the SH coefficients $ \vec{z} $, and $ \omega_k $ is the $ k $-th Fourier coefficient for the density or the specific SH coefficient.
Rendering remains completely differentiable and the time-dependent densities and SH coefficients can be reconstructed using the IDFT:
\begin{equation}\label{eq:IDFT}
x(t) = \sum_{k=0}^{K - 1} \omega_k \cdot \textrm{IDFT}_k(t)
\end{equation}
with
\begin{equation}
\textrm{IDFT}_k(t) = \begin{cases}
    \cos\left(\frac{k \pi}{T} t\right) & \textrm{if $k$ is even,}\\
    \sin\left(\frac{(k + 1) \pi}{T} t\right) & \textrm{if $k$ is odd.}
\end{cases}
\end{equation}

%% file: sections/method.tex
\section{FPO Analysis}\label{analysis}

Upon investigation of the FPO representations of a dynamic scene, we especially notice geometric reconstruction errors that are visible as ghosting artifacts of scene parts from other time steps.
While the DFT in general is able to represent arbitrary discrete sequences using $K=2T-1$ Fourier coefficients, we observe that cutting off high frequencies for the purpose of compression leads to artifacts that are equally distributed across the entire signal.
These artifacts persist even after fine-tuning which implies that the lower-dimensional representation of the signal cannot capture the crucial characteristics of the original values at each time step from the static reconstructions.
However, especially the density functions always exhibit the same properties that upon analysis lead to two key observations.

When dealing with static PlenOctrees that have been optimized independently, it is possible that leaf entries are highly varying in terms of the estimated density and color, even though the rendered results are similar. 
The reason for this effect lies in the underlying volume rendering which involves the exponential function in Eq. \ref{eq:acccolor} and \ref{eq:transmittance} to compute the observed color and transmittance based on $\sigma$.
For large input values, this function saturates, which can lead to large differences where scene content appears similarly opaque, as shown in Fig.\,\ref{fig:opacity}.

Compressing a time-dependent function with a reduced amount of frequencies in Fourier space returns a smoothed approximation of the original function.
Fig.\,\ref{fig:comp_observation} shows this effect for different settings of the number of components $K_\sigma$ kept to represent the signal.
Fewer frequencies thereby result in smoother functions and higher reconstruction error, especially visible with sharp peaks.
Density values with a higher absolute difference to the average ${ \bar{\sigma} = 1/T \sum_{t=0}^{T-1}\sigma(t) }$ are reconstructed with higher error.
This interferes with faithfully reconstructing areas with low or zero-densities, such as empty space or transparent and semi-opaque surfaces.
However, large positive values do not need to be reconstructed as precisely.
The saturation property of the transfer function allows for higher reconstruction errors of large positive values, which is visualized in Fig.\,\ref{fig:saturation_observation}.
The reconstruction with only the compressed IDFT exhibits large errors, whereas after applying the transfer function, high densities are still approximated well.
Scaling down the range of values by applying for instance a logarithmic function automatically allows for a higher approximation error of high densities after the inverse transformation.

In addition to the compressed IDFT, the $L_2$-minimization of the fine-tuning process treats approximation errors of all input values equally.
This is not necessary for large densities in opaque areas and leads to the conclusion that the density reconstruction needs to be concentrated on low and zero-density areas.

During rendering an FPO, zero-densities are interpreted as free space and color computations are skipped for these locations.
Negative densities generally cannot be interpreted in a meaningful way.
However, during rendering and fine-tuning, colors only need to be evaluated for existing geometry, which is represented with positive densities.
Negative values are ignored and can be interpreted to represent free space.
Thus, an implicit clipping via the ReLU function lifts the restriction that free space has to be represented as a zero-value.
With this observation, we can grant more freedom to the representation of zero-density values and, thus, also to the DFT approximation.

\section{Density Encoding} \label{sec:method}

\begin{figure}
    \centering
    \begin{minipage}{0.47\textwidth}
         \centering
         \includegraphics[width=\textwidth]{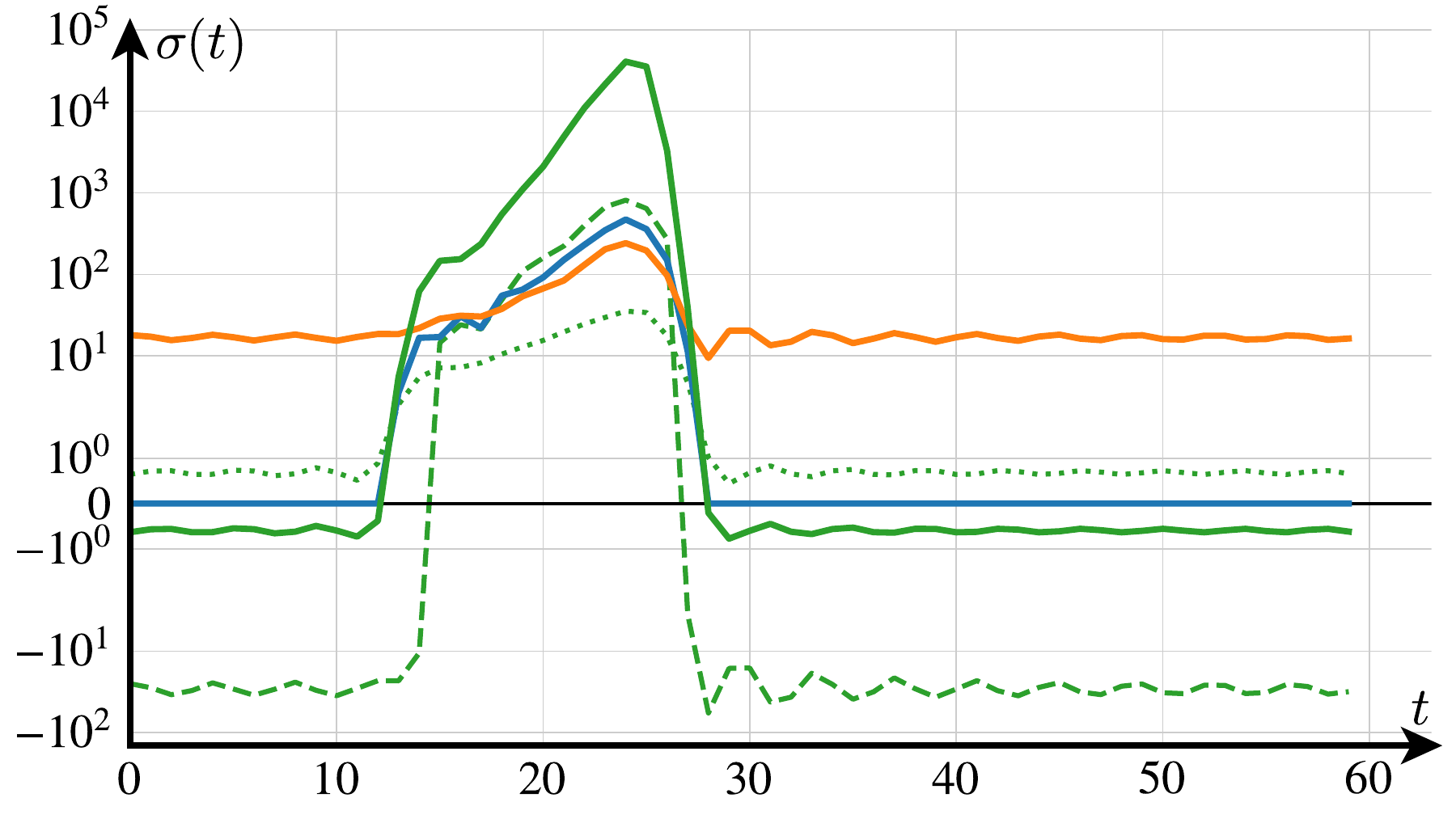}
     \end{minipage}\\
     \begin{minipage}{0.4\textwidth}
         \centering
         \includegraphics[width=\textwidth]{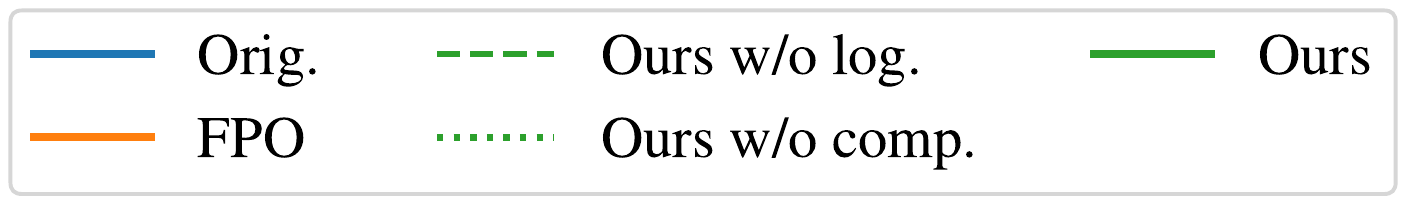}
     \end{minipage}
    \caption{Reconstruction of a density function (Orig.) using only DFT and IDFT as proposed for FPOs~\cite{wang2022fourier} and additionally in combination with our component-dependent (comp.) and logarithmic (log.) encoding on top of the DFT and IDFT.}
    \label{fig:scaling}
\end{figure}

Based on the insights of the aforementioned analysis, we propose an encoding for the densities to facilitate the reconstruction of the original $\sigma$.
During the FPO construction, we perform the compression on encoded densities 
\begin{equation}
    \sigma' = e_{\text{comp}}(e_{\log}(\sigma))
\end{equation}
where the encoding consists of a component-dependent and a logarithmic part.
We use the latter also during rendering and fine-tuning the FPO, while we apply the former only as an initialization during construction.

Fig.\,\ref{fig:scaling} shows the differences in the reconstruction of a density function using both or only one part of the encoding.
The encoding allows for a better reconstruction of the original densities without any fine-tuning than can be achieved with only the DFT and IDFT.

\subsection{Logarithmic Density Encoding} \label{logscaling} 

We use the observation that high density values can have a larger approximation error without impairing the rendered result to improve the reconstruction with the IDFT.
To weigh the values according to their importance in reconstruction and focus the approximation on densities near or equal to zero, larger values should be mapped closer together, while smaller values should stay almost the same.

This property is satisfied when encoding the individual non-negative density values $\sigma$ logarithmically using
\begin{equation} \label{eq:log_fnc}
    e_{\log}(\sigma) = \log(\sigma +1)
\end{equation}
before applying the DFT.
We choose the shift by 1 so that $e_{\log}$ remains a non-negative function for non-negative input densities with $e_{\log}(0) = 0$.
During rendering, we apply the inverse of Eq. \ref{eq:log_fnc} after the IDFT to project the densities back to their original range.
The effect of this encoding can be seen in Fig.\,\ref{fig:scaling}.

The encoded density sequences are easier to approximate with a low-frequency Fourier basis exploiting the properties of the transfer function.
Furthermore, the $L_2$-error in fine-tuning is allowed to be larger for encoded high values than without logarithmic encoding, and we focus the optimization on the more important parts of the reconstruction.

\subsection{Component-dependent Encoding} \label{componentscaling}

With the DFT, an approximation of the original function is reconstructed, where low $\sigma$ values are increased, while high $\sigma$ values are reduced.
This leads to an under-estimation of its variations over time.

Intuitively, using fewer components leads to this under-estimation as fewer frequencies are summed up to reconstruct the original function.
The heights of the peaks in the function are correlated with the ratio
\begin{equation}
    s(K_\sigma)= 0.5 \cdot (K_\sigma+1) / T
\end{equation}
between the number of frequencies $ K_\sigma $ and the number of time steps $ T $, as shown in Fig.\,\ref{fig:comp_observation}.
Amplitudes of the density function are, however, not smaller compared to zero but relative to the average $\bar{\sigma}$.
Smaller than average values thus need to be reduced further.

We shift the densities by $\sigma_{\text{shift}}$ and then scale them with the inverse ratio $1 / s(K_\sigma)$ before the DFT during FPO construction for a better approximation:
\begin{align}
    & e_{\text{comp}}(\sigma) = \frac{1}{s(K_\sigma)} \cdot (\sigma - \sigma_{\text{shift}}) + \sigma_{\text{shift}}
    \\
    & \sigma_{\text{shift}} = 
        \begin{cases}
            \bar{\sigma} & \text{if } \exists\,t \in\{0,\dots,T-1\} \colon \sigma(t) = 0, \\
            0 & \text{otherwise.}
        \end{cases}
\end{align}
In octree leaves that only contain positive $\sigma$ for all $t$, applying $e_{\text{comp}}$ with a shift by the mean value $\bar{\sigma}$ can lead to undesired non-positive values, where positive $\sigma$ can be accidentally pushed below zero and cause holes in the reconstructed model.
Thus, the shifting is only applied if empty space and, in turn, at least one zero-density is encountered.
While such non-positive $\sigma$ may still be introduced to the reconstruction, most cases that would lead to significant errors in the reconstruction can be handled faithfully this way.

This scaling can lead to higher amplitudes in the reconstruction than desired.
However, this is not problematic following the two key observations: both large and small values can be larger or smaller, respectively, to achieve the same result.
We can largely remove geometric artifacts from incorrectly reconstructed zero-values using this technique.
The effect of the scaling is shown in Fig.\,\ref{fig:scaling}.
We apply $e_{\text{comp}}$ only during FPO construction for a better initialization of the Fourier components in the octree leaves, so its inverse and the involved values of $\bar{\sigma}$ do not have to be used and stored for rendering or fine-tuning.
Since the component-dependent encoding introduces negative densities, it needs to be applied after $e_{\log}$.

%% file: sections/results.tex
\section{Experimental Results}

\subsection{Datasets}
We use synthetic data sets of the \textit{Lego} scene from the NeRF synthetic data set~\cite{mildenhall2020nerf} and of a walking human model (\textit{Walk}) generated from motion data from the CMU Motion Capture Database~\cite{cmumotion2022} using a human model~\cite{makehuman2022}.
Each data set includes 125 inward-facing camera views with a resolution of ${800\,\times\,800}$ pixels per time step anchored to the model, where $20\,\%$ are used for testing purposes.
The real-world NHR data set consisting of four scenes (\textit{Basketball}, \textit{Sport\,1}, \textit{Sport\,2} and \textit{Sport\,3}) including corresponding masks \cite{wu2020multi} are used for evaluation on real scenes.
\textit{Basketball} includes 72 views of ${1024\,\times\,768}$ and ${1224\,\times\,1024}$ resolution, where 7 views are withheld for testing purposes, whereas the \textit{Sport} data sets each contain 56 views with 6 views withheld for testing.

\subsection{Training}

Since the reference implementation of the original Fourier PlenOctrees~\cite{wang2022fourier} is unfortunately not publically available and also relies on a generalizable NeRF~\cite{wang2021ibrnet} that has been fine-tuned on the commercial Twindom data set~\cite{twindom}, we evaluate our approach against a reimplementation, which is further referred to as FPO-NGP.
In particular, we employed Instant-NGP~\cite{mueller2022instant} instead of a generalizable NeRF and further removed floater artifacts~\cite{wirth2023post}.

\begin{figure*}[t]
    \centering
    \begin{minipage}{0.03\textwidth}
        \centering \hspace{0.5cm}
    \end{minipage}\hfill
    \begin{minipage}{0.31\textwidth}
        \centering Lego
    \end{minipage}\hfill
    \begin{minipage}{0.21\textwidth}
        \centering Walk
    \end{minipage}\hfill
    \begin{minipage}{0.19\textwidth}
        \centering Basketball
    \end{minipage}\hfill
    \begin{minipage}{0.25\textwidth}
        \centering Sport 1
    \end{minipage}\\
    \begin{minipage}{0.03\textwidth}
        \centering
        \begin{sideways}
            FPO-NGP
        \end{sideways}
    \end{minipage}\hfill
    \begin{minipage}{0.31\textwidth}
        \centering
        \includegraphics[width=0.48\textwidth,trim={48 90 60 100},clip]{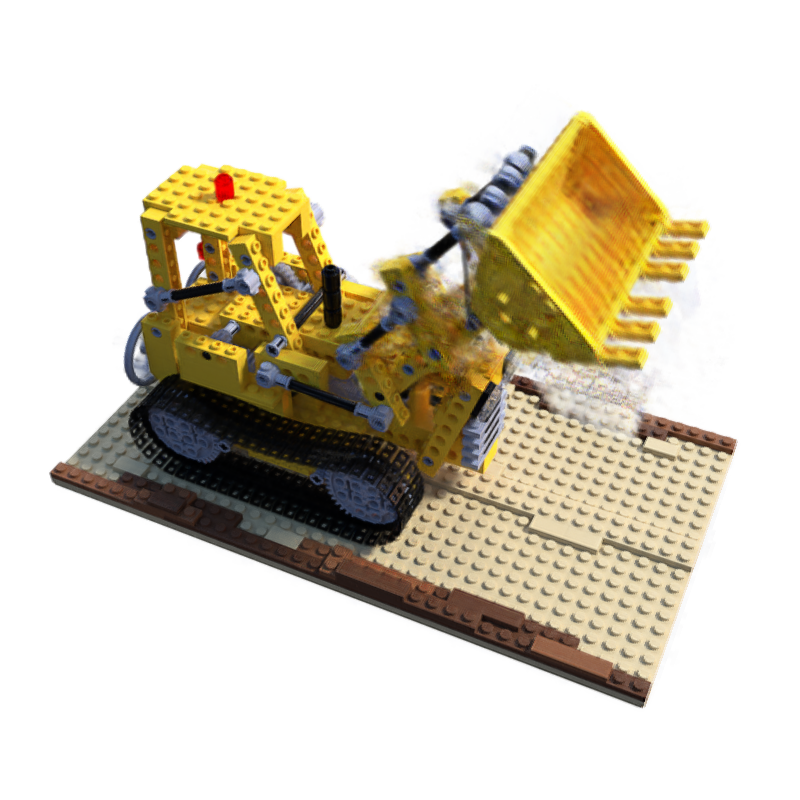}
        \includegraphics[width=0.48\textwidth,trim={20 50 100 100},clip]{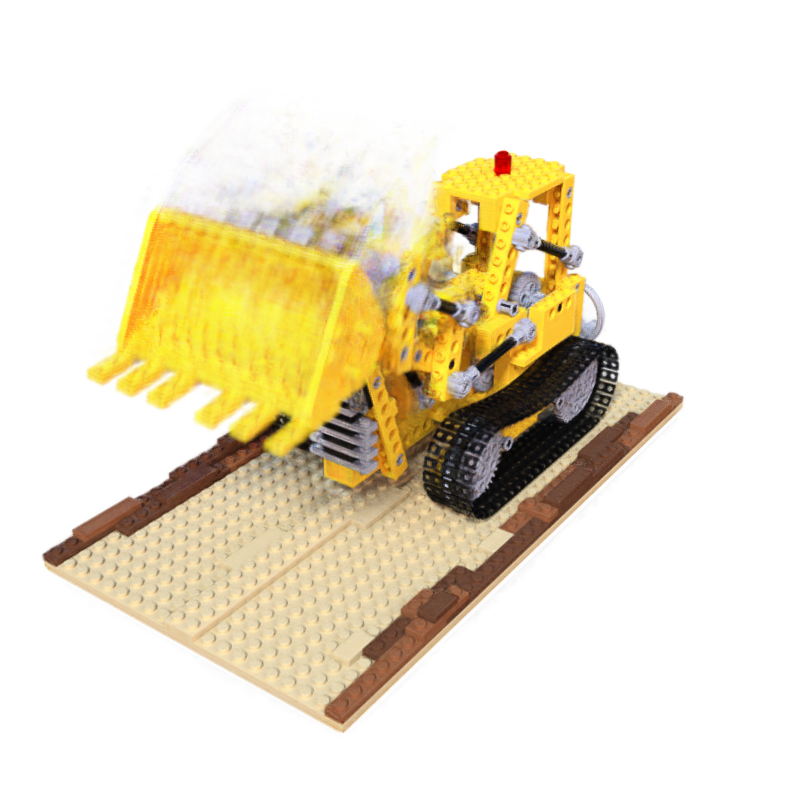}
    \end{minipage}\hfill
    \begin{minipage}{0.21\textwidth}
        \centering \vspace{0.5cm}
        \includegraphics[width=0.49\textwidth,trim={230 70 205 20},clip]{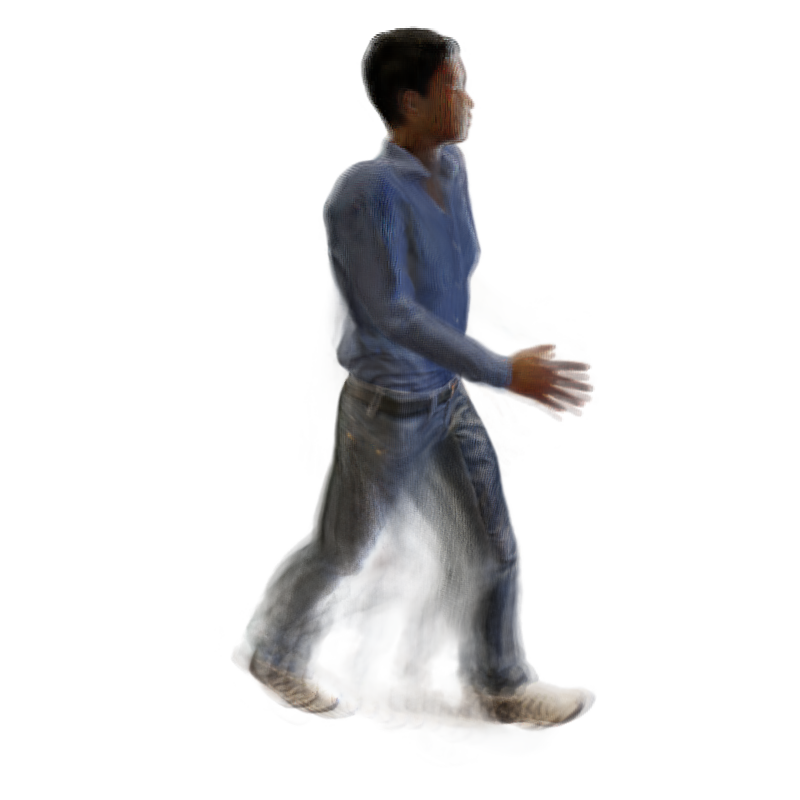}
        \includegraphics[width=0.46\textwidth,trim={240 70 220 20},clip]{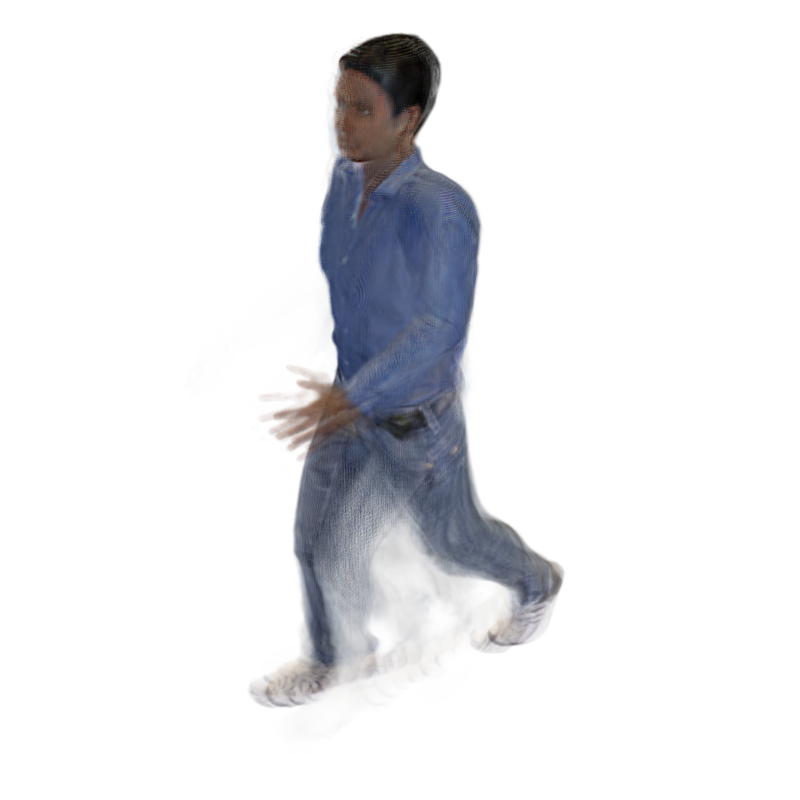}
    \end{minipage}\hfill
    \begin{minipage}{0.19\textwidth}
        \centering
        \includegraphics[width=0.46\textwidth,trim={190 30 380 230},clip]{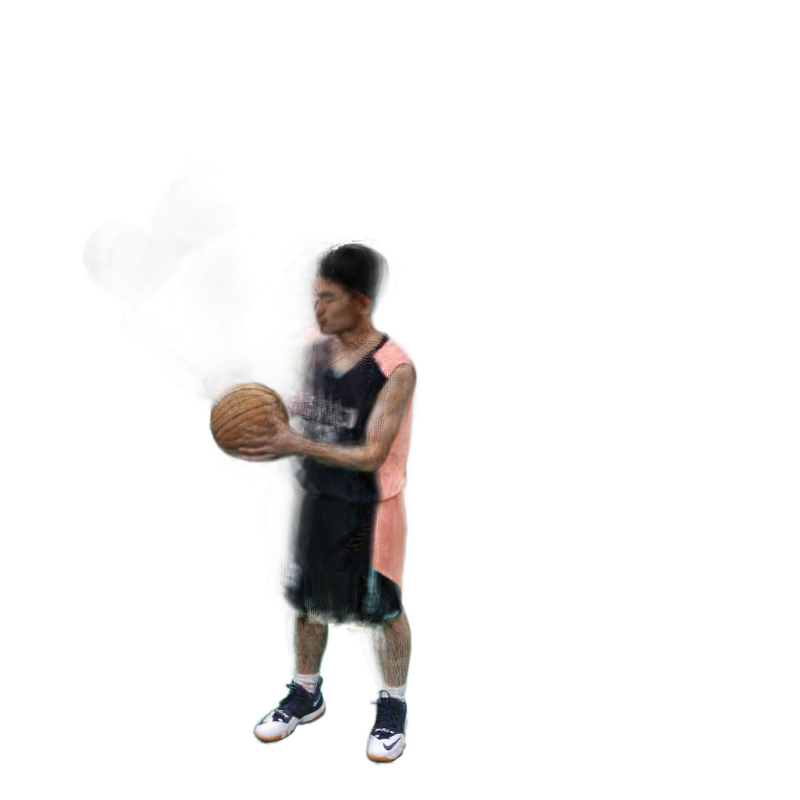}
        \includegraphics[width=0.49\textwidth,trim={330 20 220 150},clip]{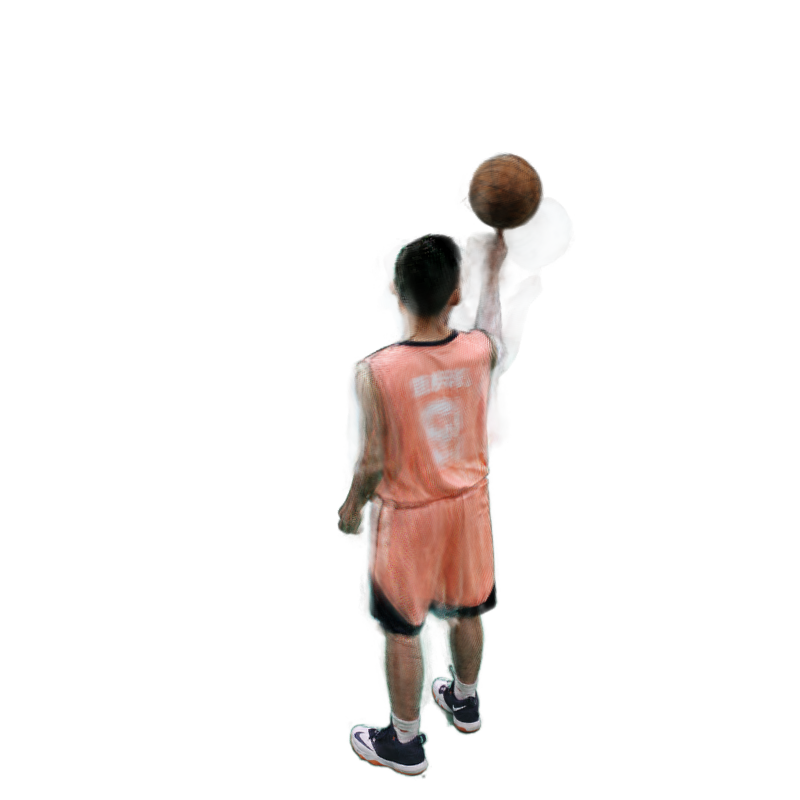}
    \end{minipage}\hfill
    \begin{minipage}{0.25\textwidth}
        \centering
        \includegraphics[width=0.33\textwidth,trim={300 85 300 230},clip]{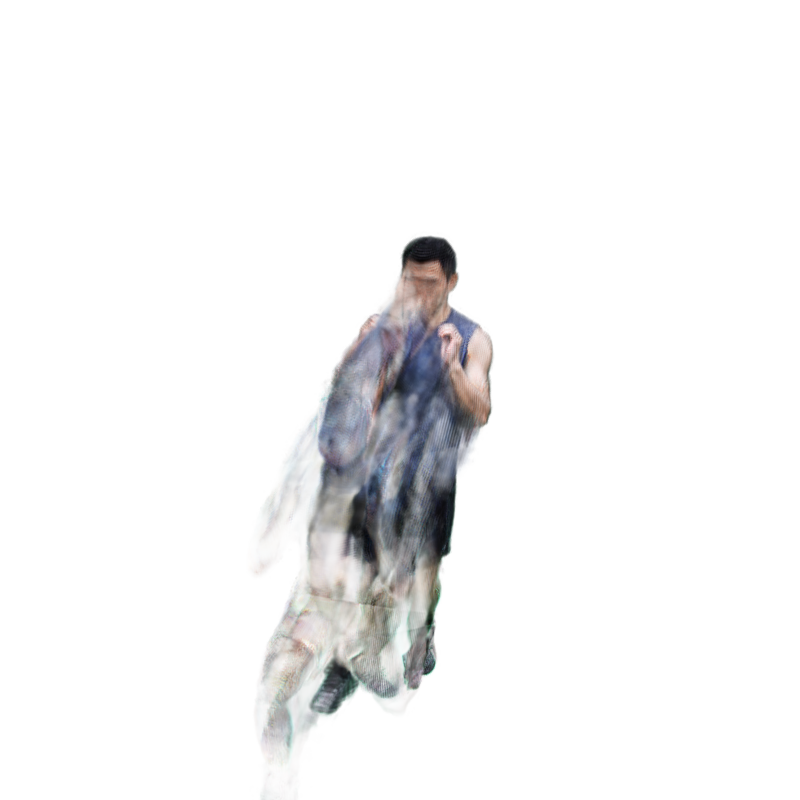}
        \includegraphics[width=0.63\textwidth,trim={110 200 230 0},clip]{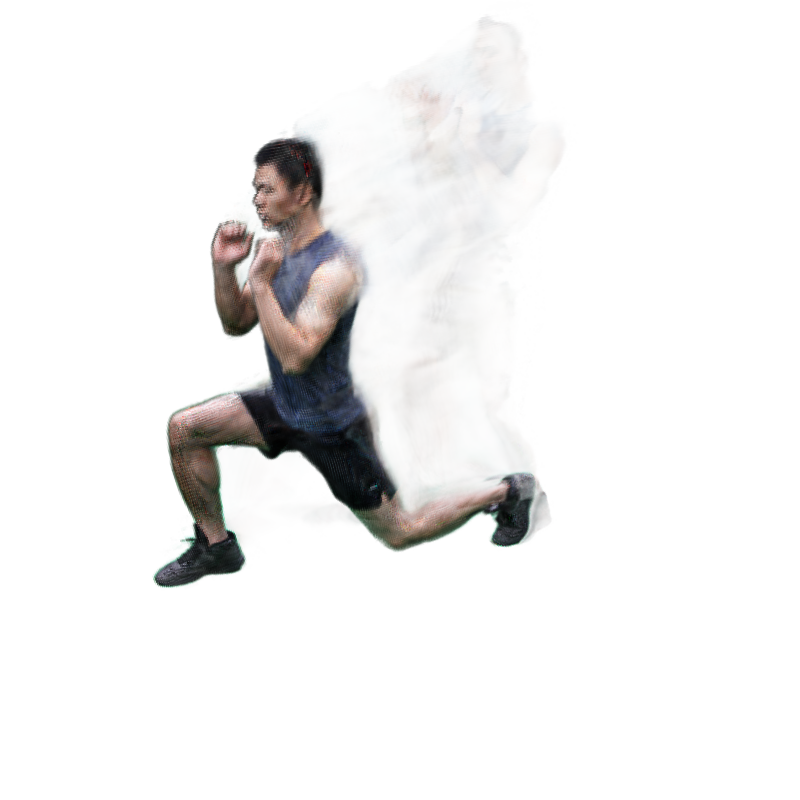}
    \end{minipage}\\
    \begin{minipage}{0.03\textwidth}
        \centering
        \begin{sideways}
            Ours
        \end{sideways}
    \end{minipage}\hfill
    \begin{minipage}{0.31\textwidth}
        \centering
        \includegraphics[width=0.48\textwidth,trim={48 90 60 100},clip]{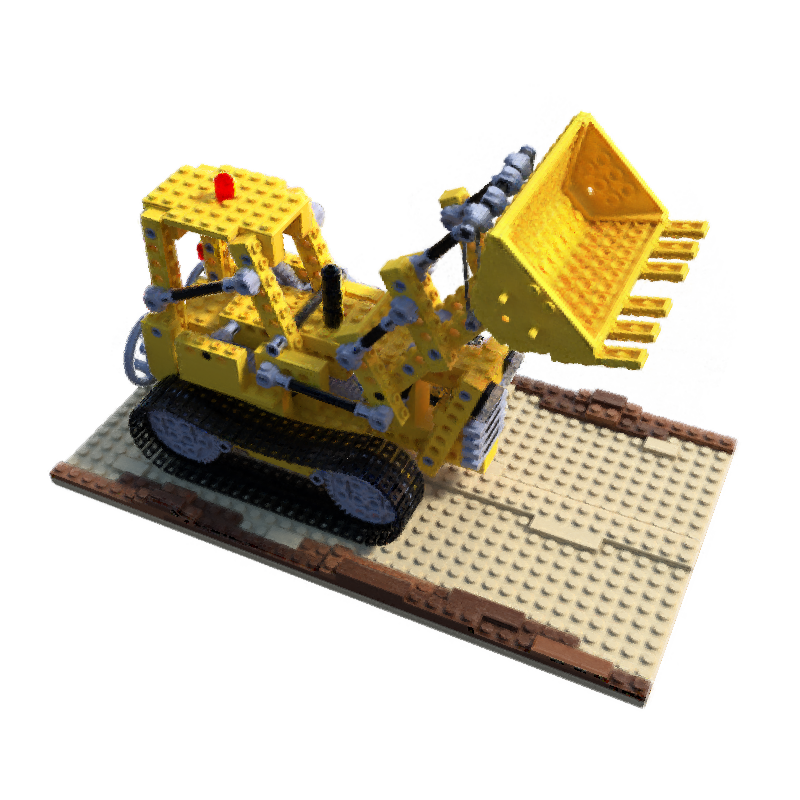}
        \includegraphics[width=0.48\textwidth,trim={20 50 100 100},clip]{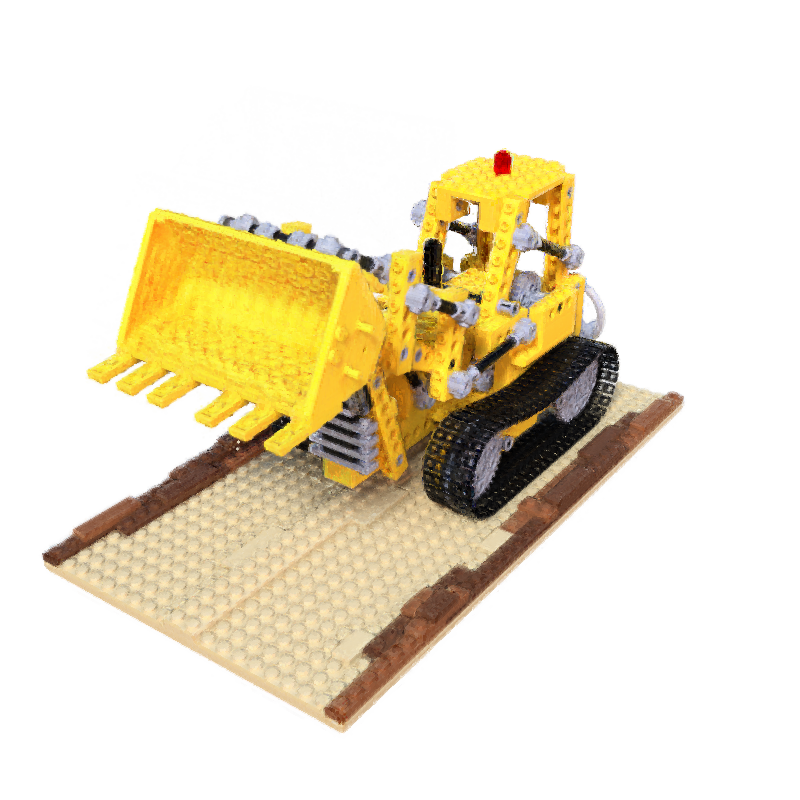}
    \end{minipage}\hfill
    \begin{minipage}{0.21\textwidth}
        \centering \vspace{0.5cm}
        \includegraphics[width=0.49\textwidth,trim={230 70 205 20},clip]{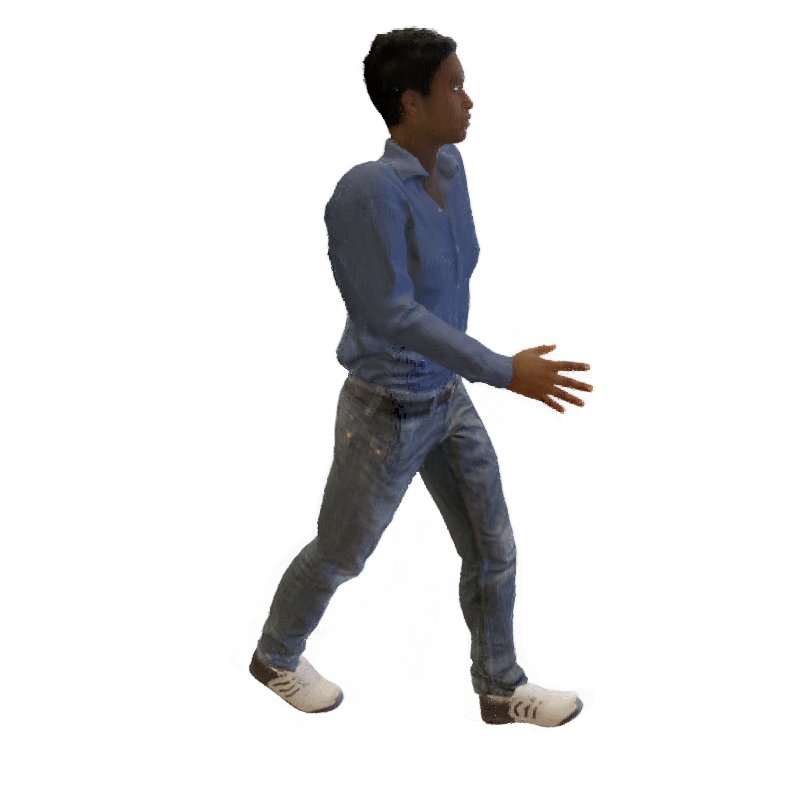}
        \includegraphics[width=0.46\textwidth,trim={240 70 220 20},clip]{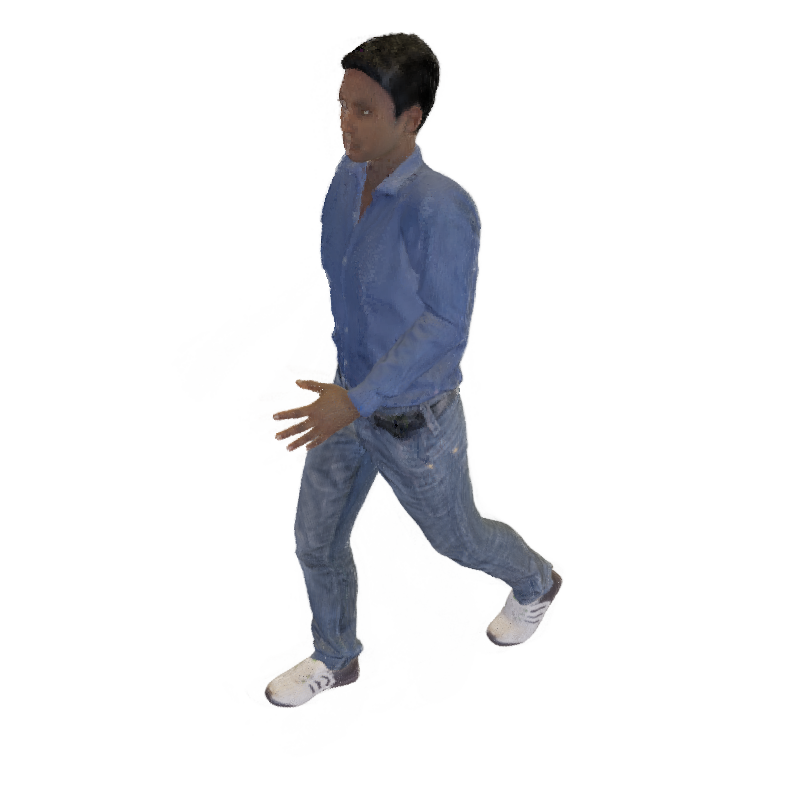}
    \end{minipage}\hfill
    \begin{minipage}{0.19\textwidth}
        \centering
        \includegraphics[width=0.46\textwidth,trim={190 30 380 230},clip]{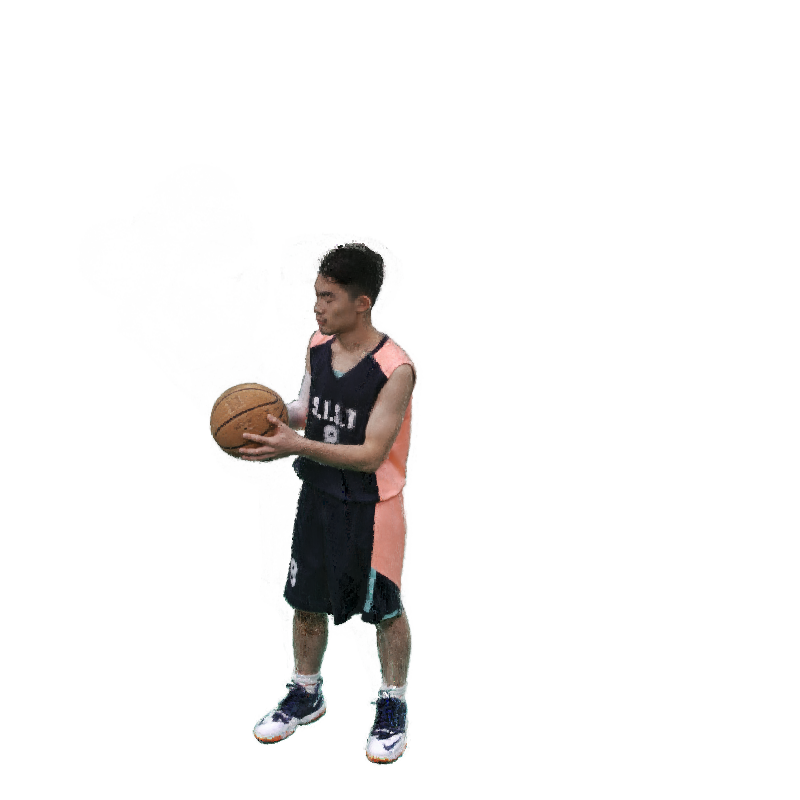}
        \includegraphics[width=0.49\textwidth,trim={330 20 220 150},clip]{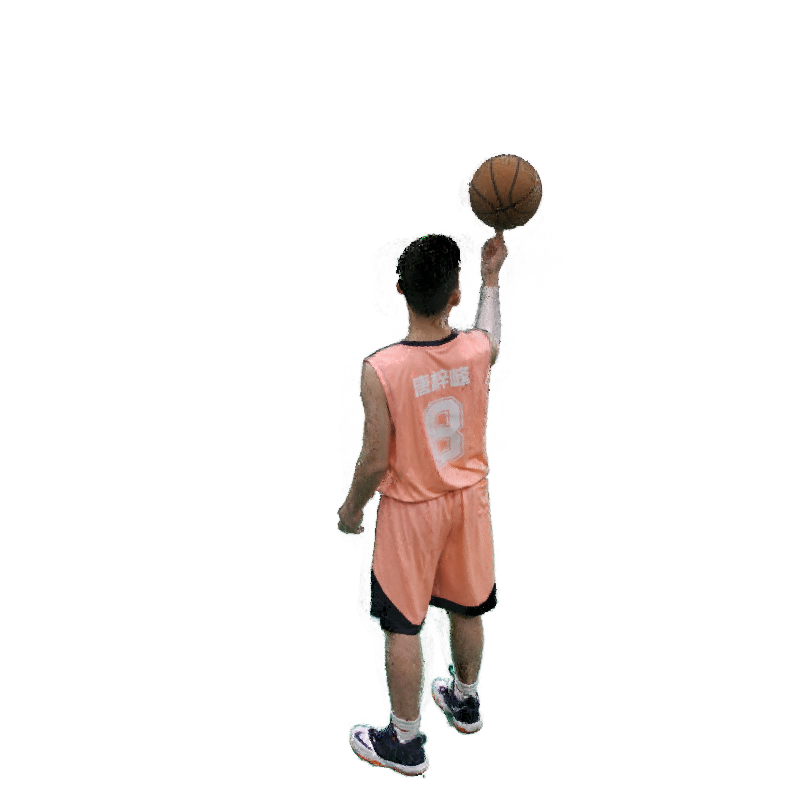}
    \end{minipage}\hfill
    \begin{minipage}{0.25\textwidth}
        \centering
        \includegraphics[width=0.33\textwidth,trim={300 85 300 230},clip]{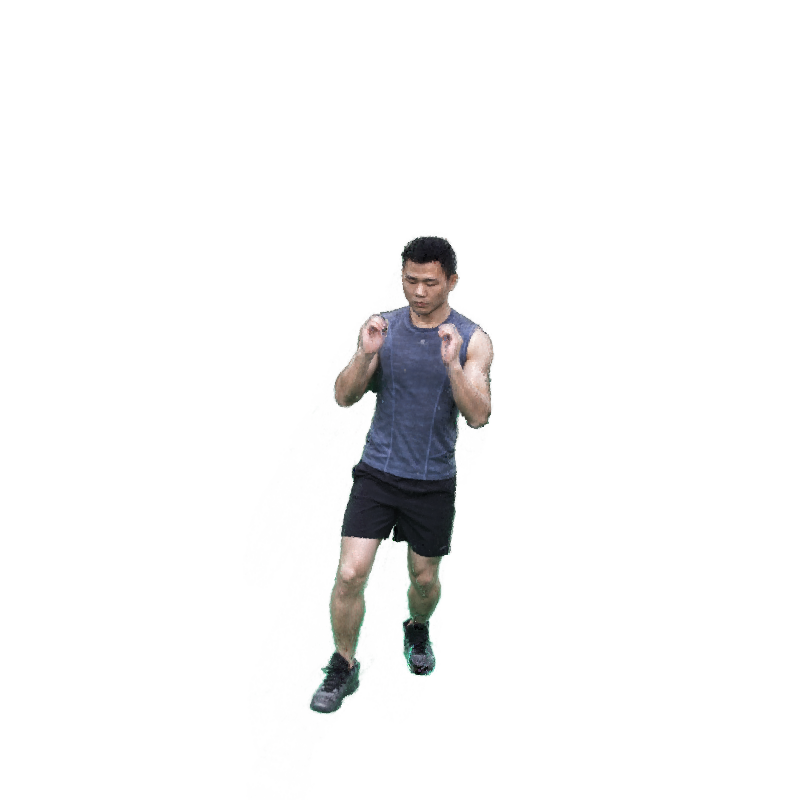}
        \includegraphics[width=0.63\textwidth,trim={110 200 230 0},clip]{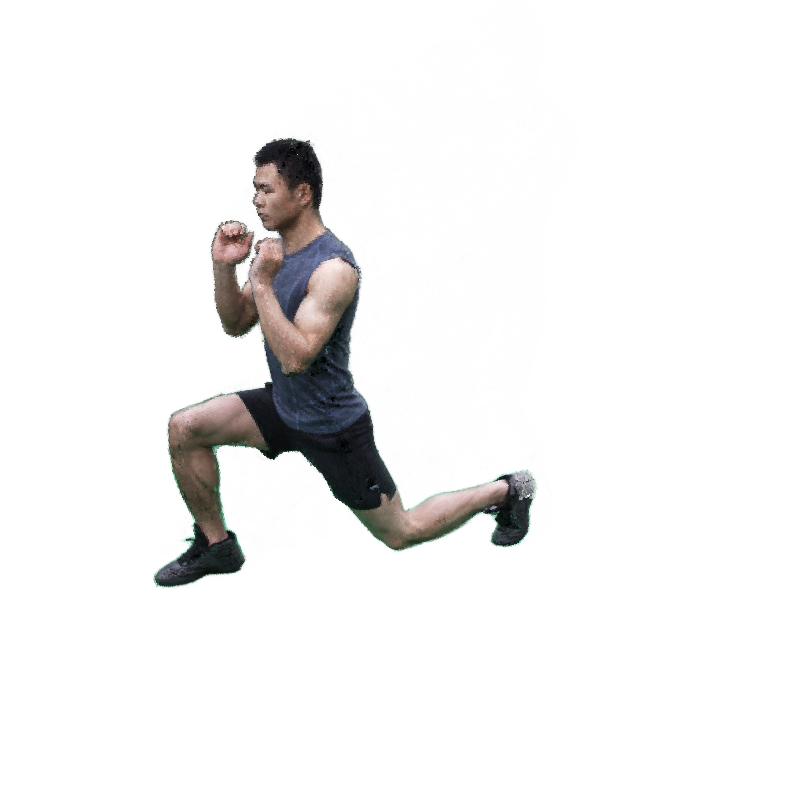}
    \end{minipage}
    \caption{Renderings of FPOs of different dynamic scenes without and with our logarithmic and component-dependent encoding after 10 epochs of fine-tuning.}
    \label{fig:example_renderings}
\end{figure*}

\begin{table*}[t]
    \centering
    \caption{Comparison of achieved metrics averaged over all data sets with different combinations of logarithmic encoding (log.) and component-dependent encoding (comp.), both before and after fine-tuning for 1 and 10 epochs. Arrows indicate whether a high value ($\uparrow$) or a low value ($\downarrow$) is better. Best and second best results are marked in green and yellow, respectively.}
    \begin{tabular}{lcccccccccccc} %ccc}
        \toprule
         Fine-tuning && \multicolumn{3}{c}{0 epochs} && \multicolumn{3}{c}{1 epoch} && \multicolumn{3}{c}{10 epochs}\\
         && PSNR$\uparrow$ & SSIM$\uparrow$ & LPIPS$\downarrow$ && PSNR$\uparrow$ & SSIM$\uparrow$ & LPIPS$\downarrow$ && PSNR$\uparrow$ & SSIM$\uparrow$ & LPIPS$\downarrow$\\
         \midrule
         FPO-NGP&& 17.44&0.891&0.173 && 21.58&0.918&0.142 && 23.79&0.922&0.125 \\
         Ours w/o comp.&& 23.05&\colorbox{best}{0.926}&0.122 && \colorbox{best}{29.64}&\colorbox{best}{0.950}&\colorbox{best}{0.096} && \colorbox{best}{30.90}&\colorbox{best}{0.958}&\colorbox{best}{0.089} \\
         Ours w/o log.&& \colorbox{second}{23.68}&\colorbox{second}{0.917}&\colorbox{second}{0.121} && 26.32&0.936&\colorbox{second}{0.102} && 27.75&0.940&0.101 \\
         Ours&& \colorbox{best}{23.85}&0.910&\colorbox{best}{0.113} && \colorbox{second}{28.79}&\colorbox{second}{0.940}&\colorbox{best}{0.096} && \colorbox{second}{29.90}&\colorbox{second}{0.948}&\colorbox{second}{0.094} \\
         \bottomrule
    \end{tabular}
    \label{tab:metrics}
\end{table*}

To obtain the static PlenOctrees, a set of ${T=60}$ NeRF-models is trained first.
The networks use the multiresolution hash encoding and NeRF network architecture of Instant-NGP~\cite{mueller2022instant} but produce view-independent SH coefficients instead of view-dependent RGB colors as output, analogous to NeRF-SH~\cite{yu2021plenoctrees}.
The training images are scaled down by a factor of two.

The PlenOctrees are extracted from the trained implicit reconstructions~\cite{yu2021plenoctrees} using 9 SH coefficients per color channel on a grid of size $512^3$.
The PlenOctree bounds are set to be constant over time with varying center positions to enable representing larger motions.
Fine-tuning of the static PlenOctrees is performed for 5 epochs with training images at full resolution.

We choose the same parameters for the Fourier approximation as in the original approach~\cite{wang2022fourier}: For the density, $K_\sigma=31$ Fourier coefficients are stored in the FPO while $K_\vec{z}=5$ components are used for the SH coefficients of each color channel.
The FPO is fine-tuned for 10 epochs on the randomized training images of all time steps at full resolution.
We also augment the time sequence by duplicating the first and last frame to avoid ghosting artifacts but exclude these additional two frames from the fine-tuning and the evaluation.

All training is performed on NVIDIA GeForce RTX 3090 and RTX 4090 GPUs, where frame rates and training times are listed here for the RTX 4090 GPU.
The process to obtain a fine-tuned FPO with our un-optimized implementation takes around 6 hours for training one set of the static NeRF models and around 10 to 30 minutes per epoch of fine-tuning.
Any other steps in the training require a few minutes each.

Further details of the training procedure are provided in the supplemental material.

\subsection{Evaluation}

\begin{figure*}[t]
    \centering
    \begin{minipage}{0.04\textwidth}
        \hspace{0.1cm}
    \end{minipage}\hfill
    \begin{minipage}{0.04\textwidth}
        \hspace{0.1cm}
    \end{minipage}\hfill
    \begin{minipage}{0.18\textwidth}
        \centering
        FPO-NGP
    \end{minipage}\hfill
    \begin{minipage}{0.18\textwidth}
        \centering
        Ours w/o comp.
    \end{minipage}\hfill
    \begin{minipage}{0.18\textwidth}
        \centering
        Ours w/o log.
    \end{minipage}\hfill
    \begin{minipage}{0.18\textwidth}
        \centering
        Ours
    \end{minipage}\hfill
    \begin{minipage}{0.18\textwidth}
        \centering
        Ground truth
    \end{minipage}\\
    \begin{minipage}{0.04\textwidth}
        \begin{sideways}Lego\end{sideways}
    \end{minipage}\hfill
    \begin{minipage}{0.04\textwidth}
        \begin{sideways}10 epochs \hspace{0.5cm} 0 epochs\end{sideways}
    \end{minipage}\hfill
    \begin{minipage}{0.18\textwidth}
        \centering
        \includegraphics[width=\textwidth,trim={175 450 225 100},clip]{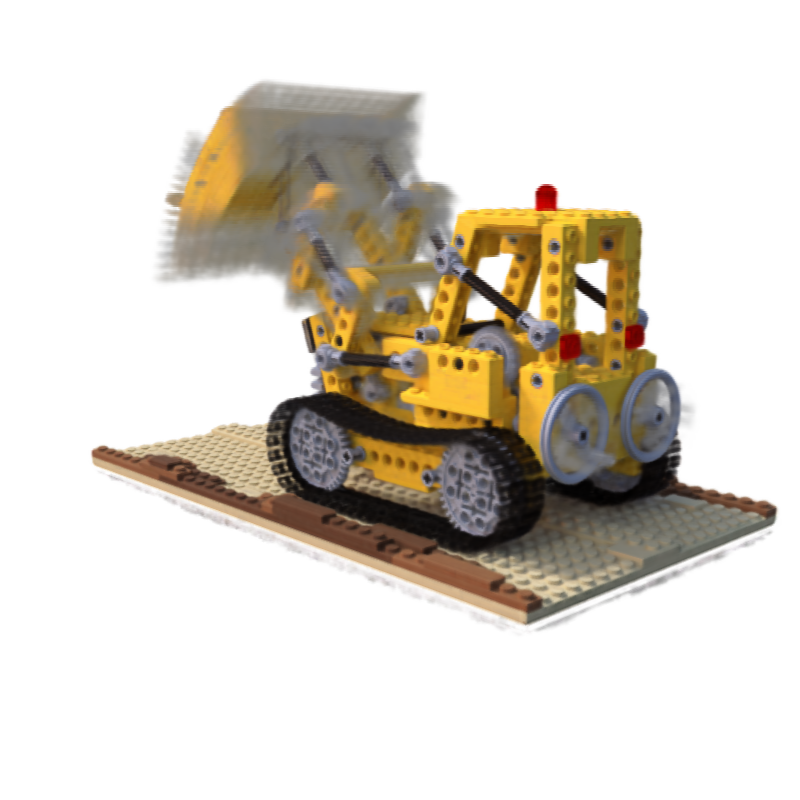}
        \includegraphics[width=\textwidth,trim={175 450 225 100},clip]{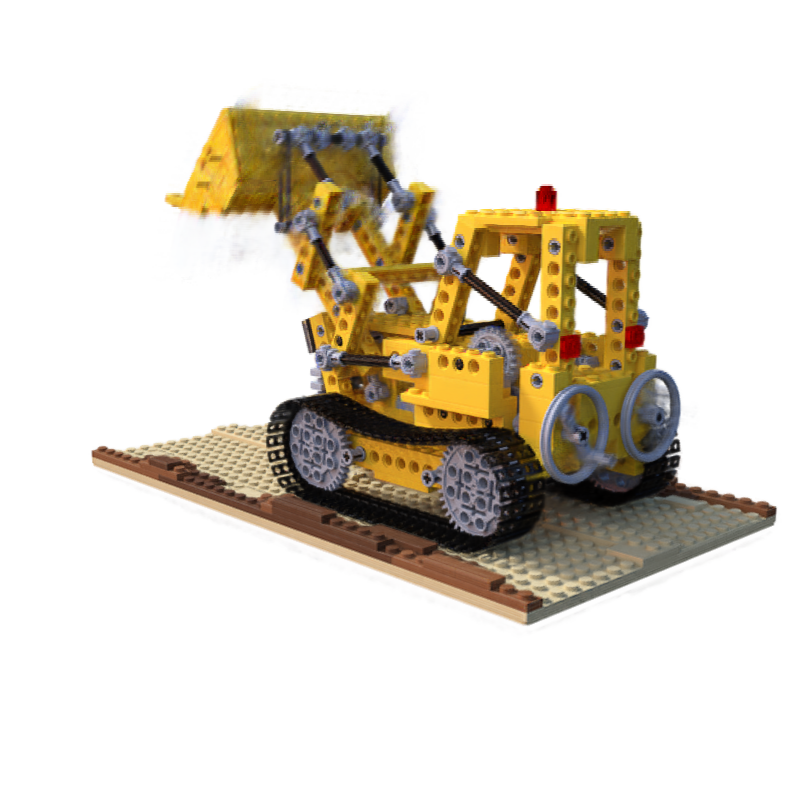}
    \end{minipage}\hfill
    \begin{minipage}{0.18\textwidth}
        \centering
        \includegraphics[width=\textwidth,trim={175 450 225 100},clip]{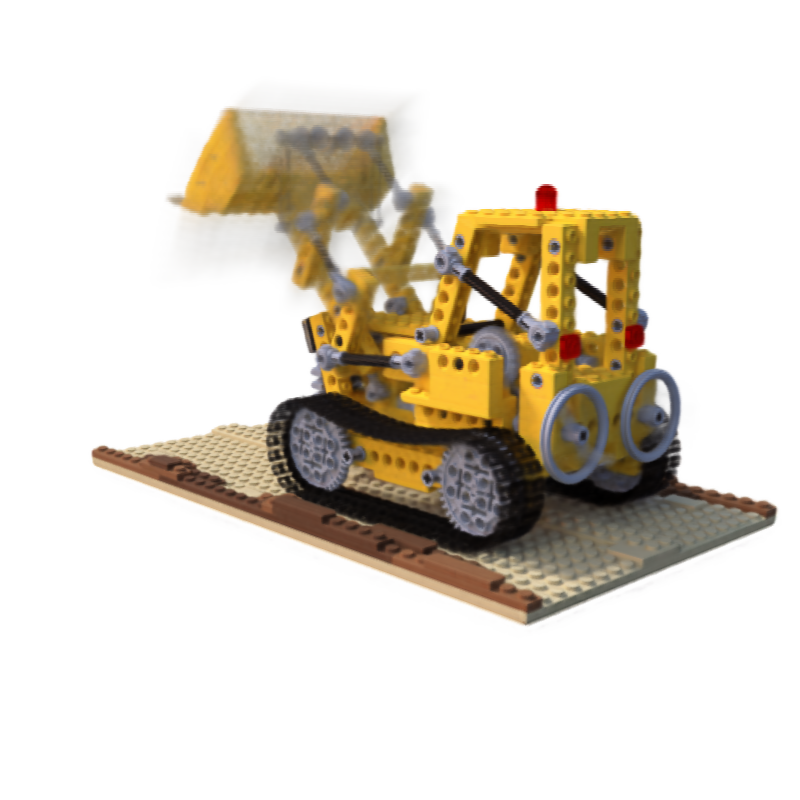}
        \includegraphics[width=\textwidth,trim={175 450 225 100},clip]{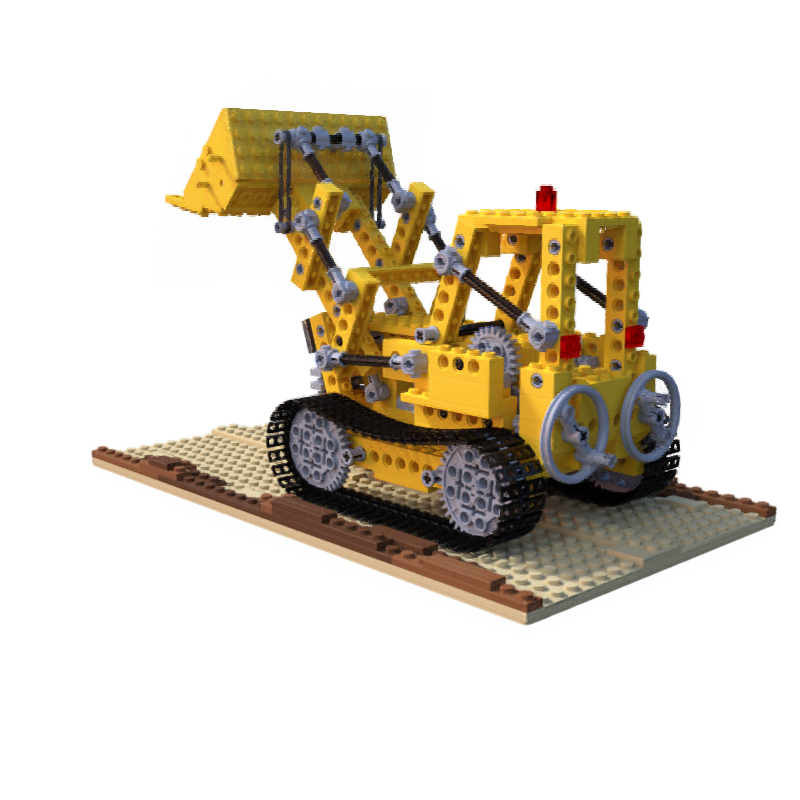}
    \end{minipage}\hfill
    \begin{minipage}{0.18\textwidth}
        \centering
        \includegraphics[width=\textwidth,trim={175 450 225 100},clip]{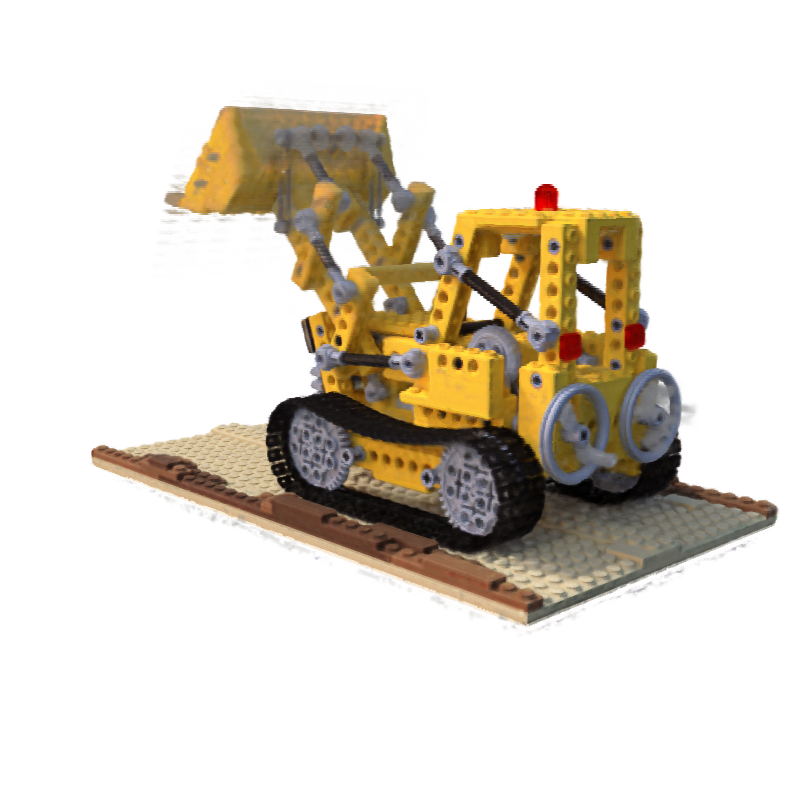}
        \includegraphics[width=\textwidth,trim={175 450 225 100},clip]{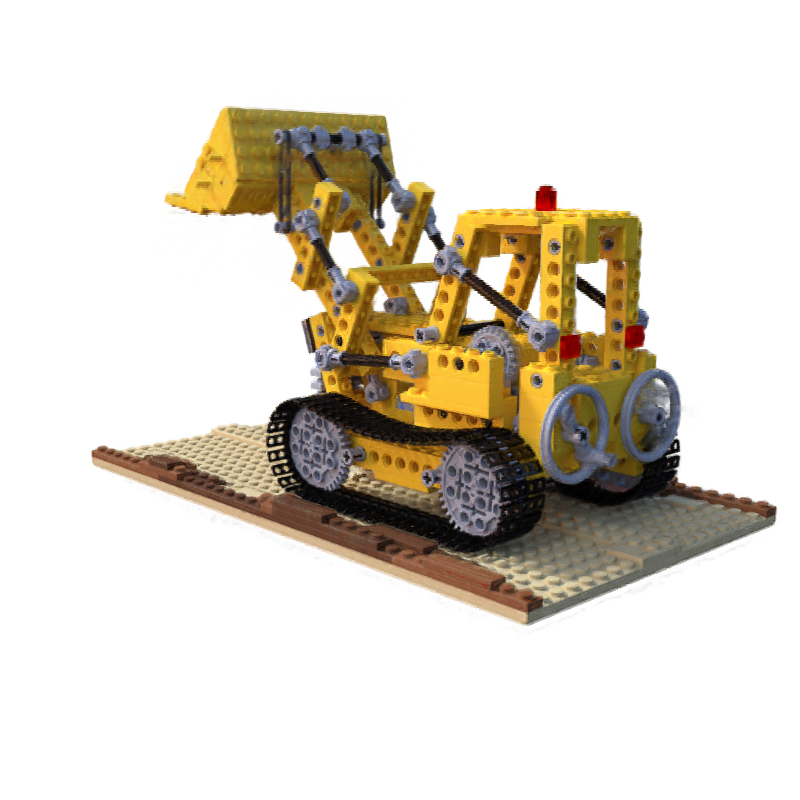}
    \end{minipage}\hfill
    \begin{minipage}{0.18\textwidth}
        \centering
        \includegraphics[width=\textwidth,trim={175 450 225 100},clip]{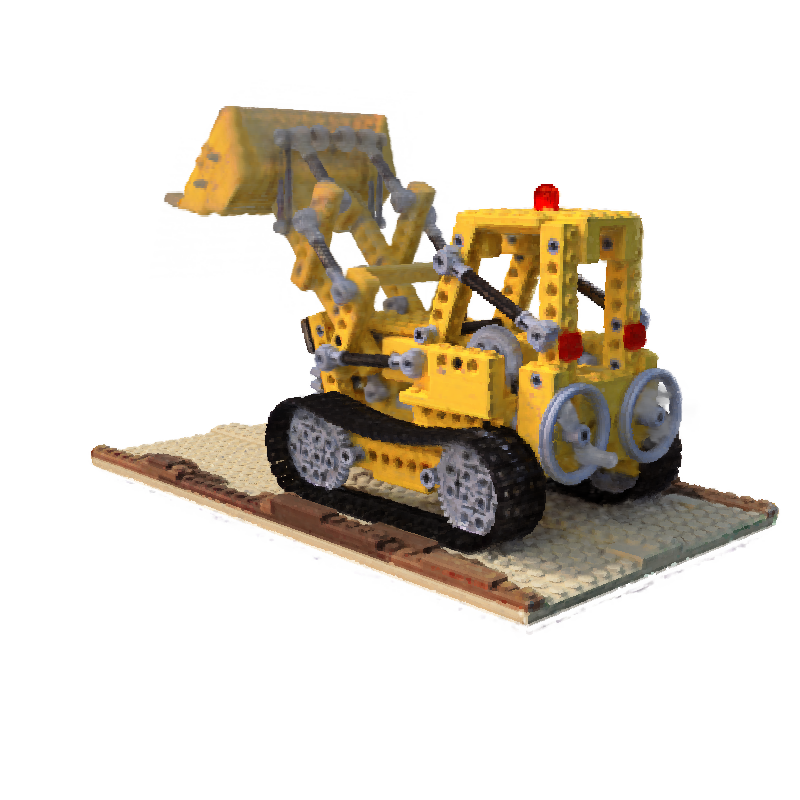}
        \includegraphics[width=\textwidth,trim={175 450 225 100},clip]{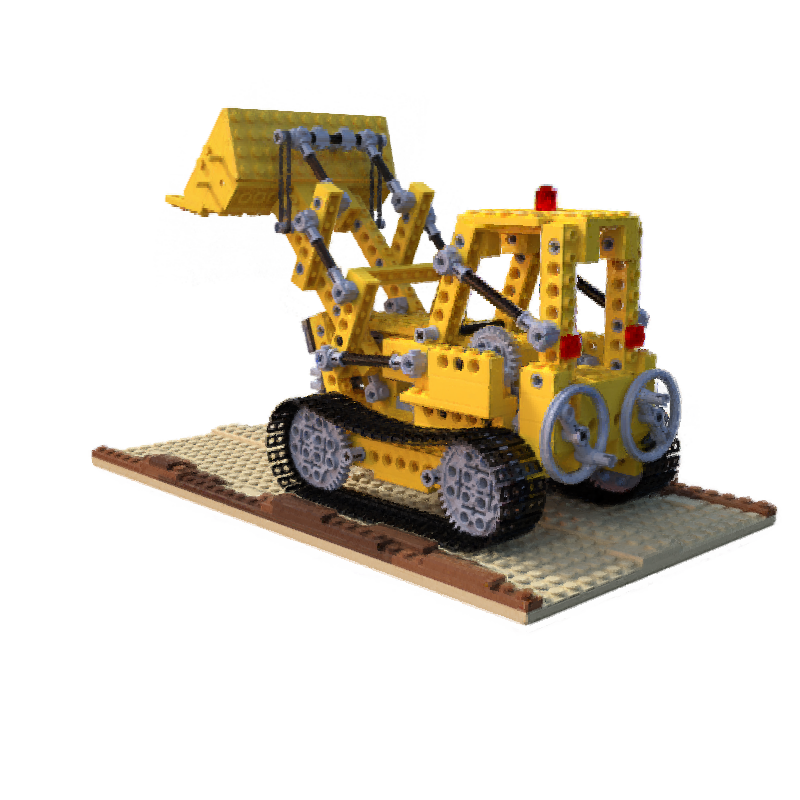}
    \end{minipage}\hfill
    \begin{minipage}{0.18\textwidth}
        \centering
        \includegraphics[width=\textwidth,trim={175 450 225 100},clip]{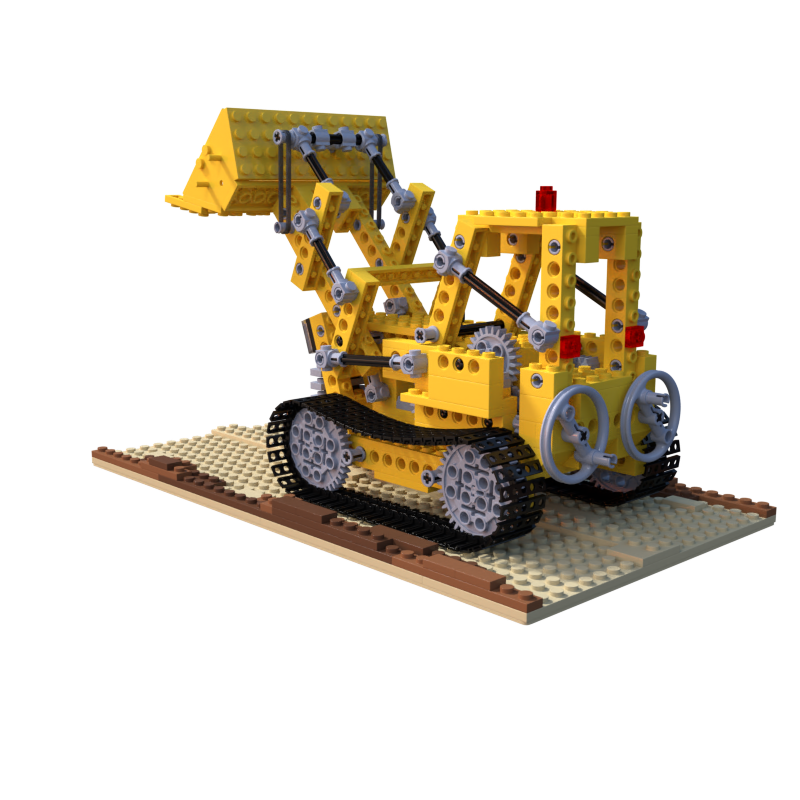}
    \end{minipage}\\
    \begin{minipage}{0.04\textwidth}
        \begin{sideways}Walk\end{sideways}
    \end{minipage}\hfill
    \begin{minipage}{0.04\textwidth}
        \begin{sideways}10 epochs \hspace{0.5cm} 0 epochs\end{sideways}
    \end{minipage}\hfill
    \begin{minipage}{0.18\textwidth}
        \centering
        \includegraphics[width=\textwidth,trim={200 175 300 437},clip]{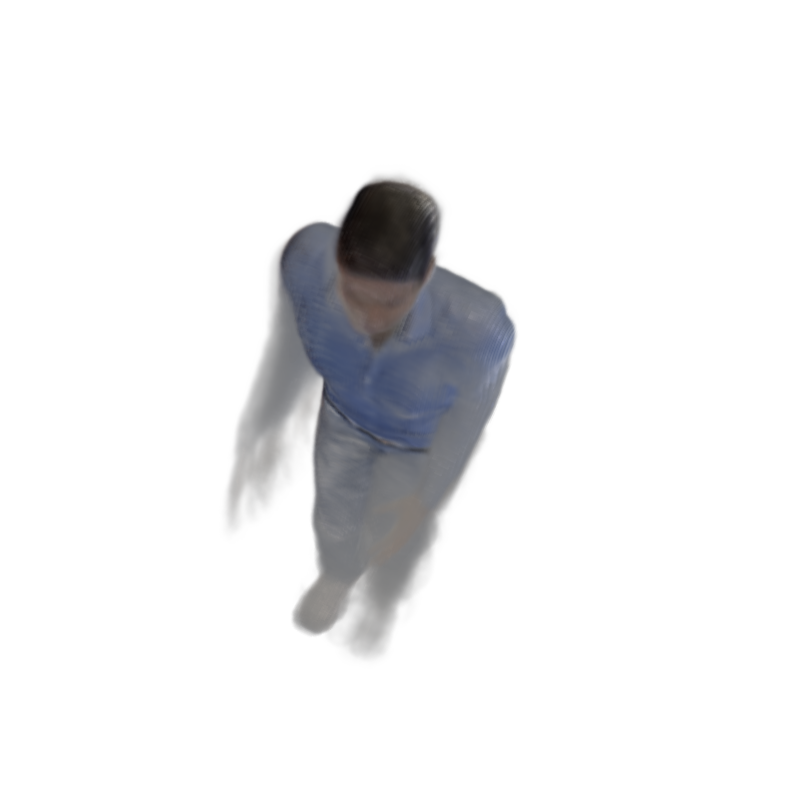}
        \includegraphics[width=\textwidth,trim={200 175 300 437},clip]{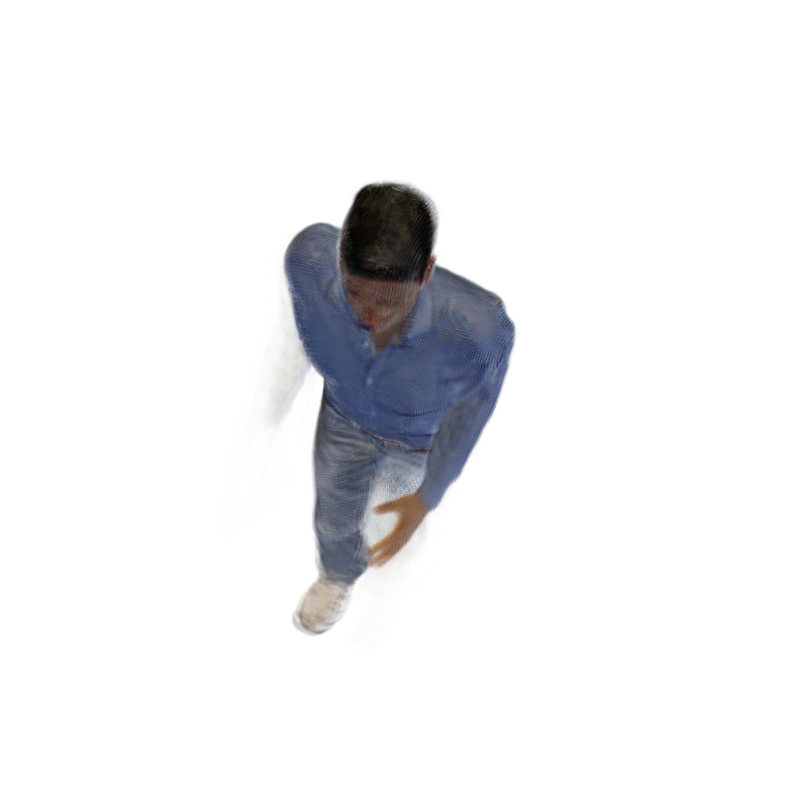}
    \end{minipage}\hfill
    \begin{minipage}{0.18\textwidth}
        \centering
        \includegraphics[width=\textwidth,trim={200 175 300 437},clip]{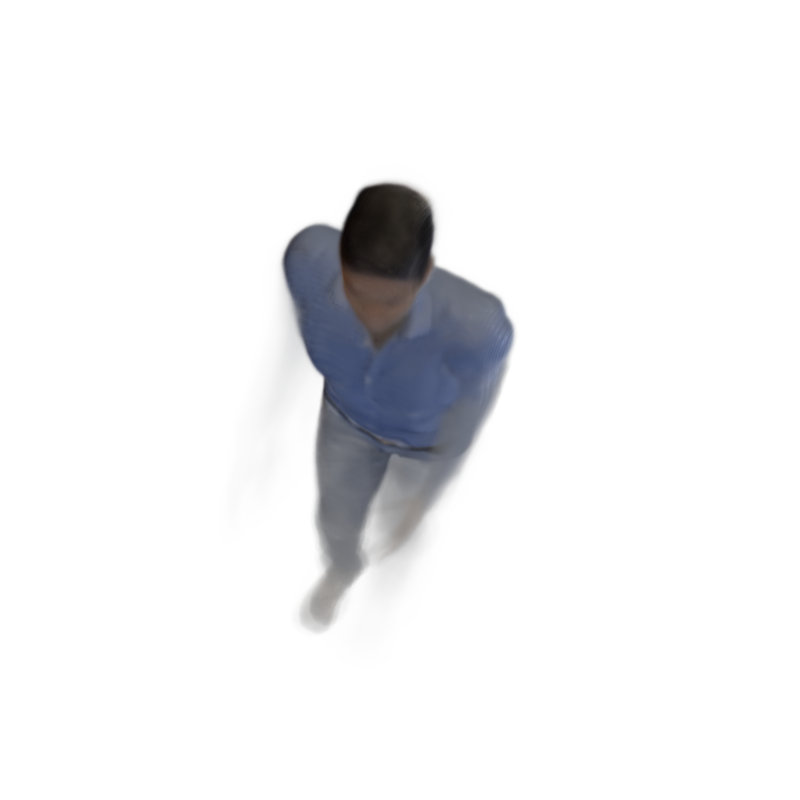}
        \includegraphics[width=\textwidth,trim={200 175 300 437},clip]{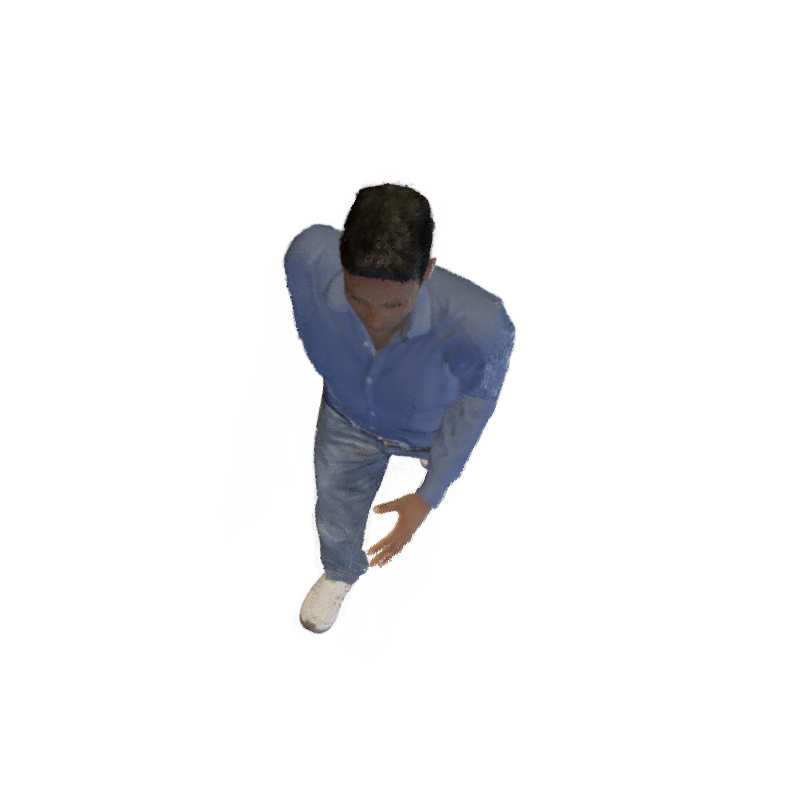}
    \end{minipage}\hfill
    \begin{minipage}{0.18\textwidth}
        \centering
        \includegraphics[width=\textwidth,trim={200 175 300 437},clip]{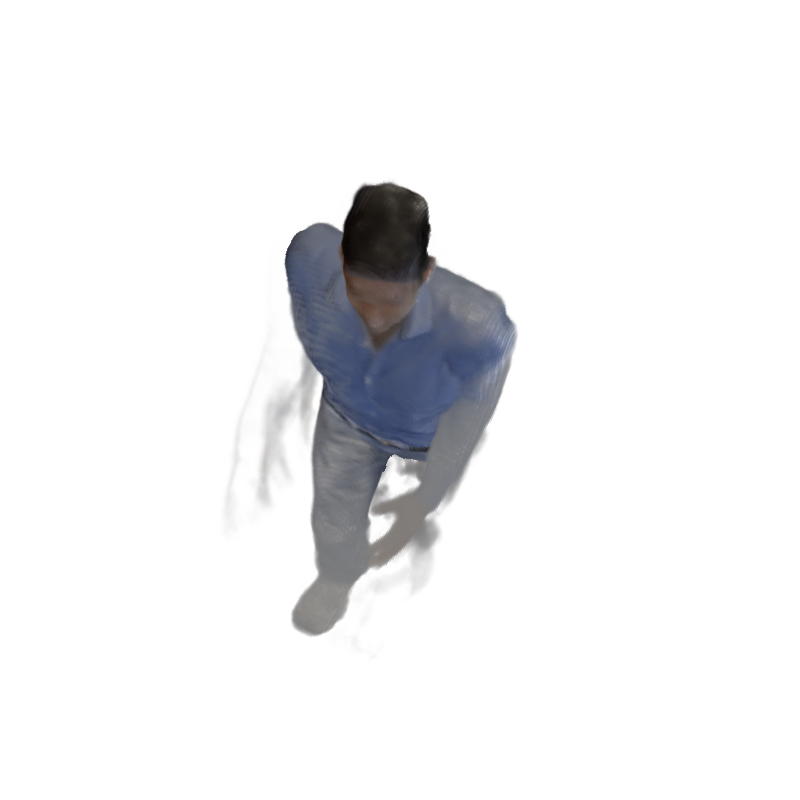}
        \includegraphics[width=\textwidth,trim={200 175 300 437},clip]{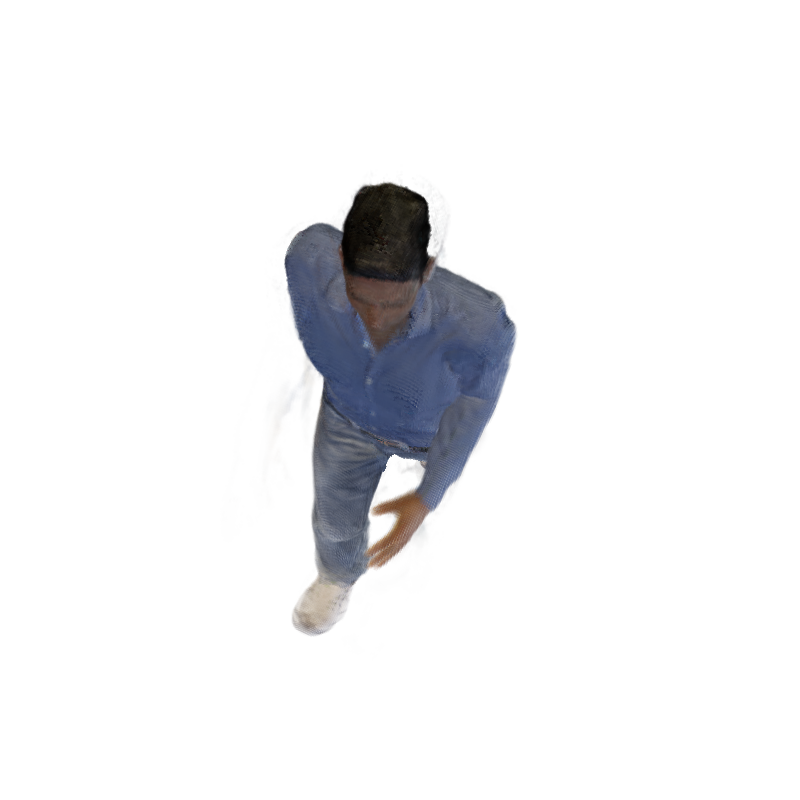}
    \end{minipage}\hfill
    \begin{minipage}{0.18\textwidth}
        \centering
        \includegraphics[width=\textwidth,trim={200 175 300 437},clip]{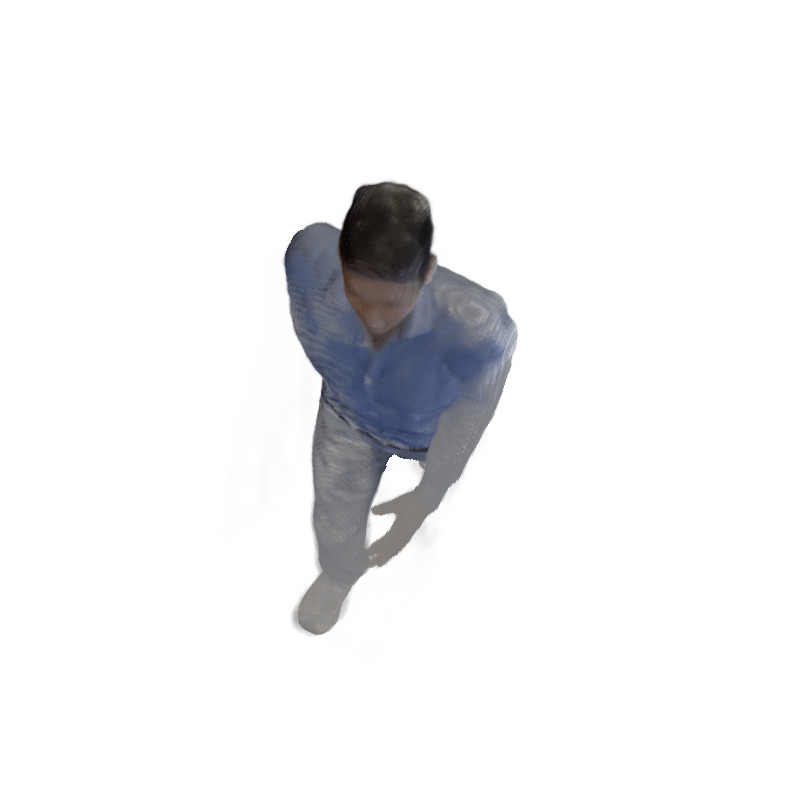}
        \includegraphics[width=\textwidth,trim={200 175 300 437},clip]{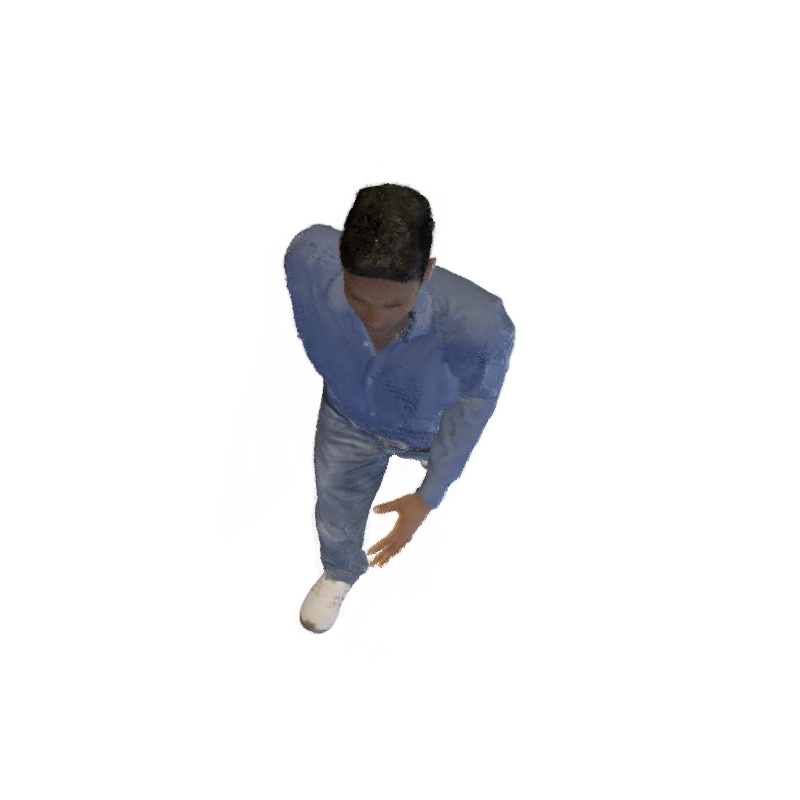}
    \end{minipage}\hfill
    \begin{minipage}{0.18\textwidth}
        \centering
        \includegraphics[width=\textwidth,trim={200 175 300 437},clip]{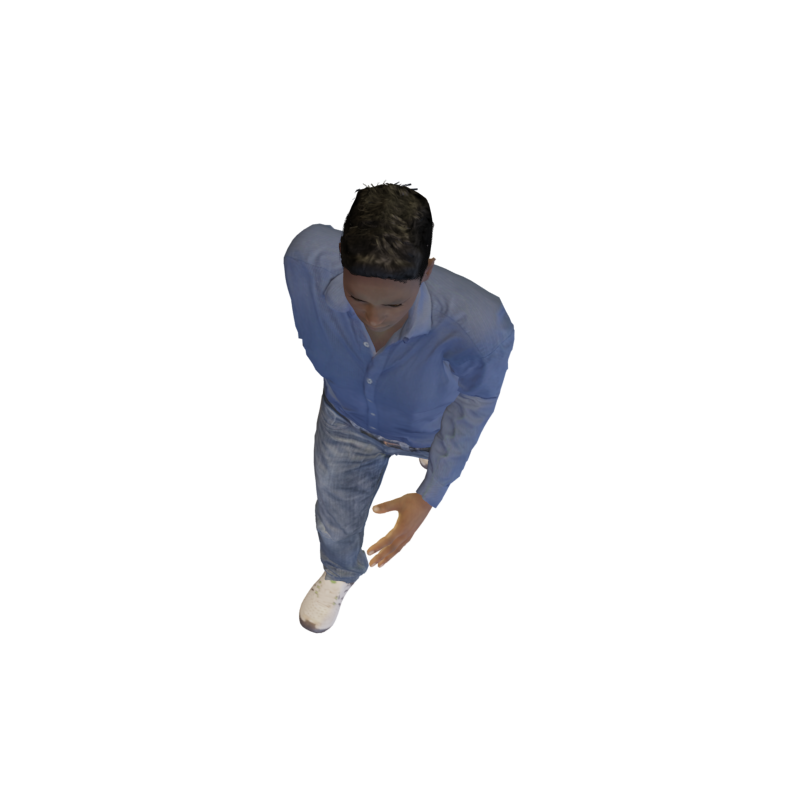}
    \end{minipage}\\
    \begin{minipage}{0.04\textwidth}
        \begin{sideways}Basketball\end{sideways}
    \end{minipage}\hfill
    \begin{minipage}{0.04\textwidth}
        \begin{sideways}10 epochs \hspace{0.5cm} 0 epochs\end{sideways}
    \end{minipage}\hfill
    \begin{minipage}{0.18\textwidth}
        \centering
        \includegraphics[width=\textwidth,trim={306 534 462 75},clip]{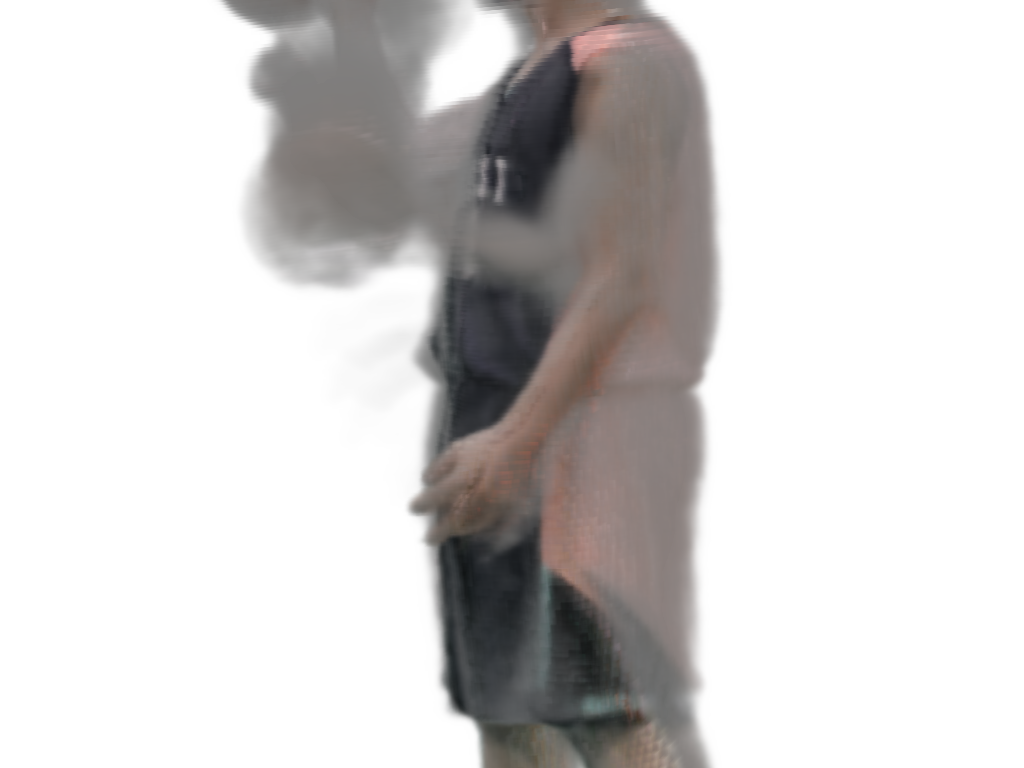}
        \includegraphics[width=\textwidth,trim={306 534 462 75},clip]{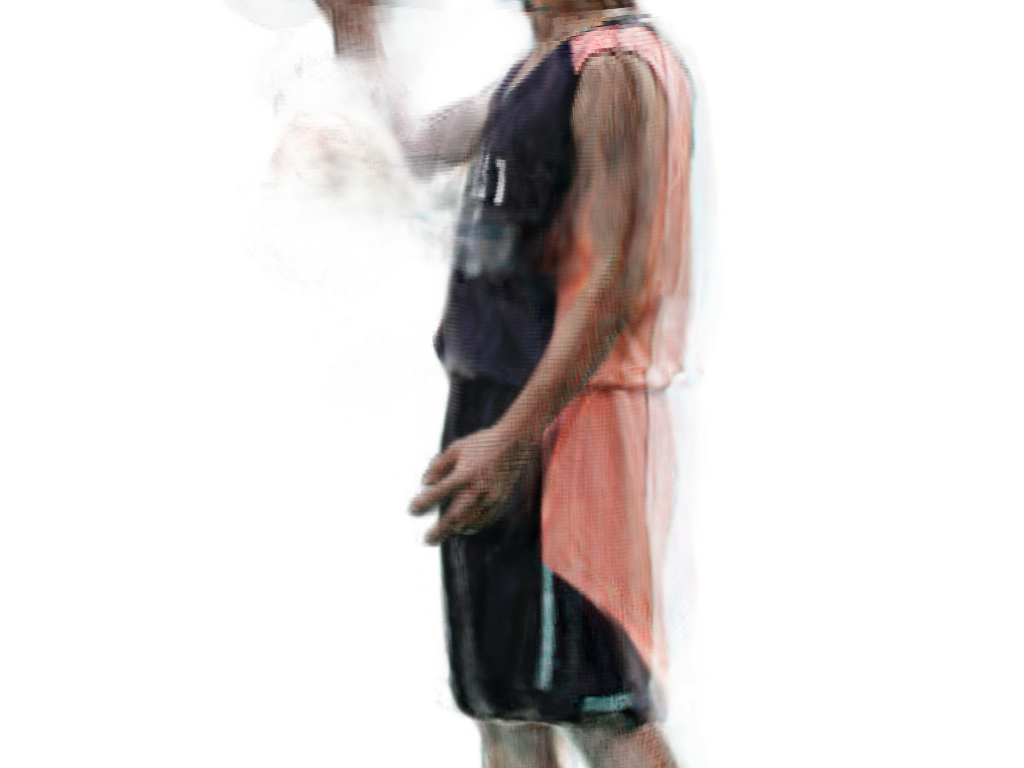}
    \end{minipage}\hfill
    \begin{minipage}{0.18\textwidth}
        \centering
        \includegraphics[width=\textwidth,trim={306 534 462 75},clip]{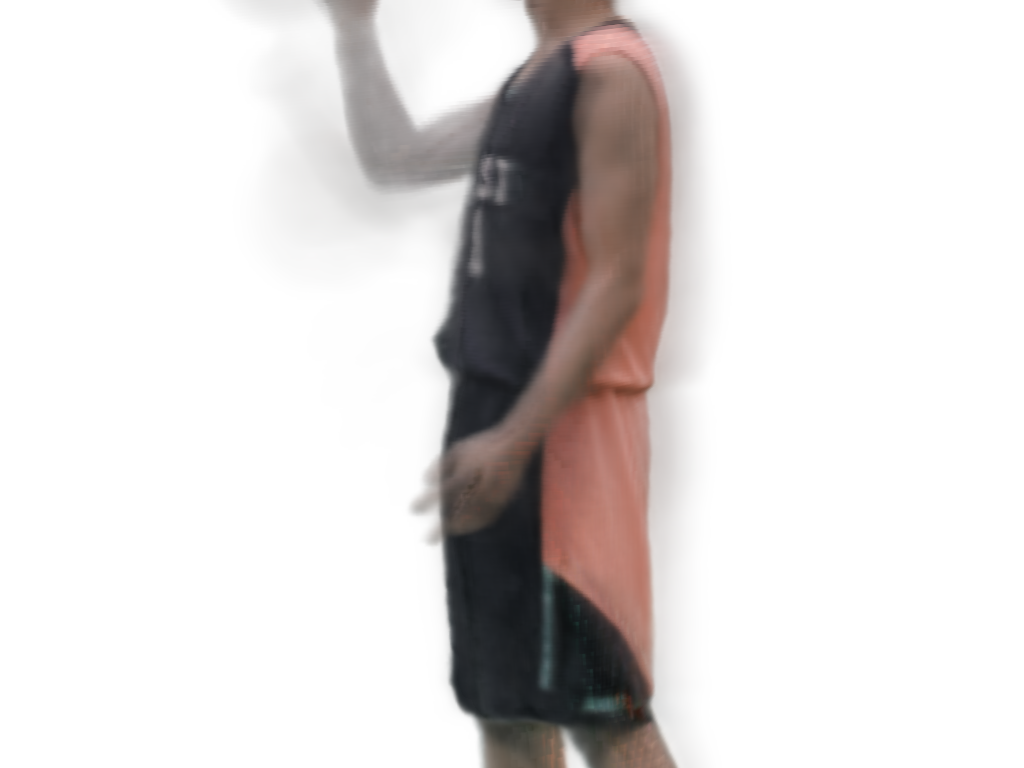}
        \includegraphics[width=\textwidth,trim={306 534 462 75},clip]{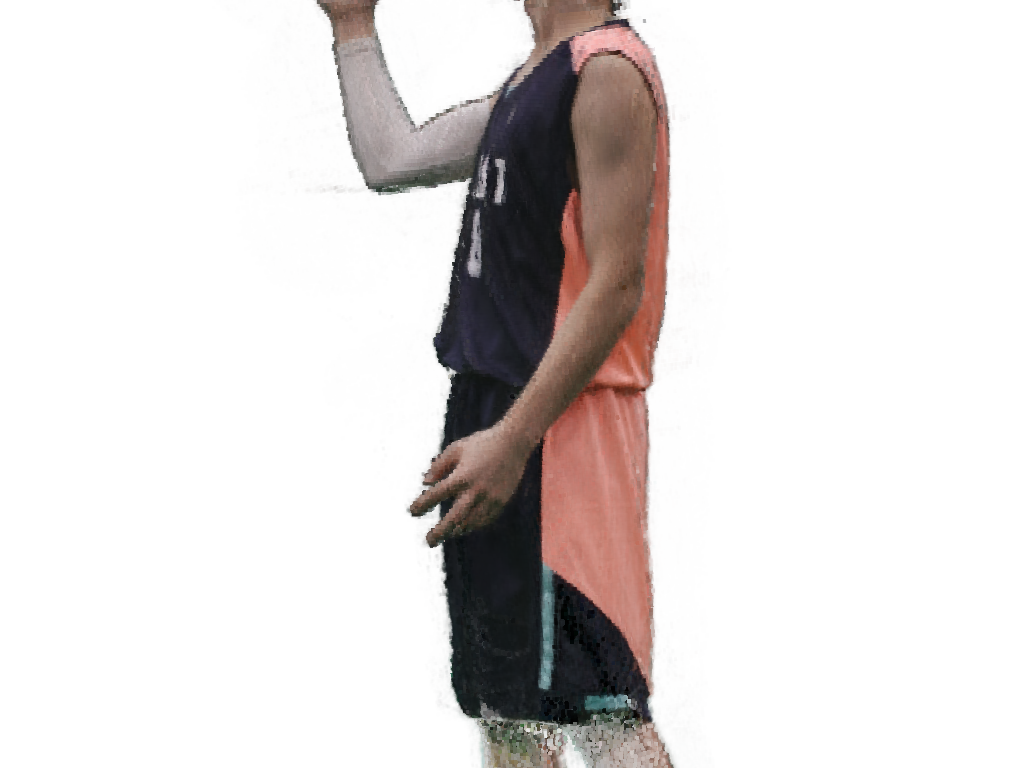}
    \end{minipage}\hfill
    \begin{minipage}{0.18\textwidth}
        \centering
        \includegraphics[width=\textwidth,trim={306 534 462 75},clip]{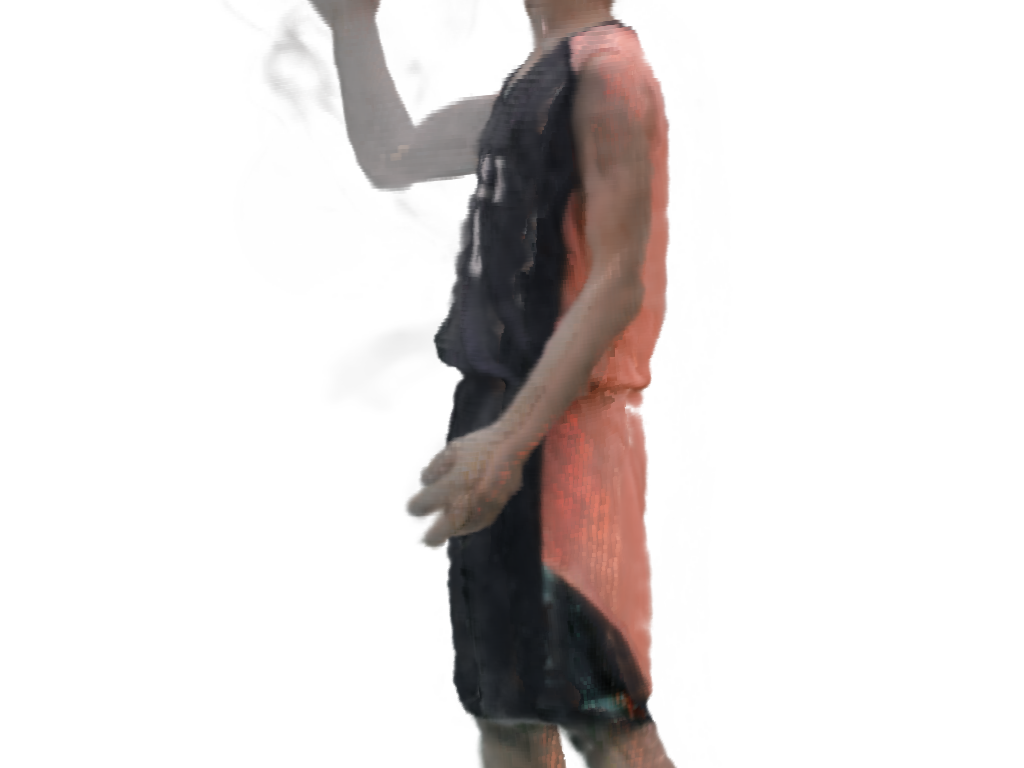}
        \includegraphics[width=\textwidth,trim={306 534 462 75},clip]{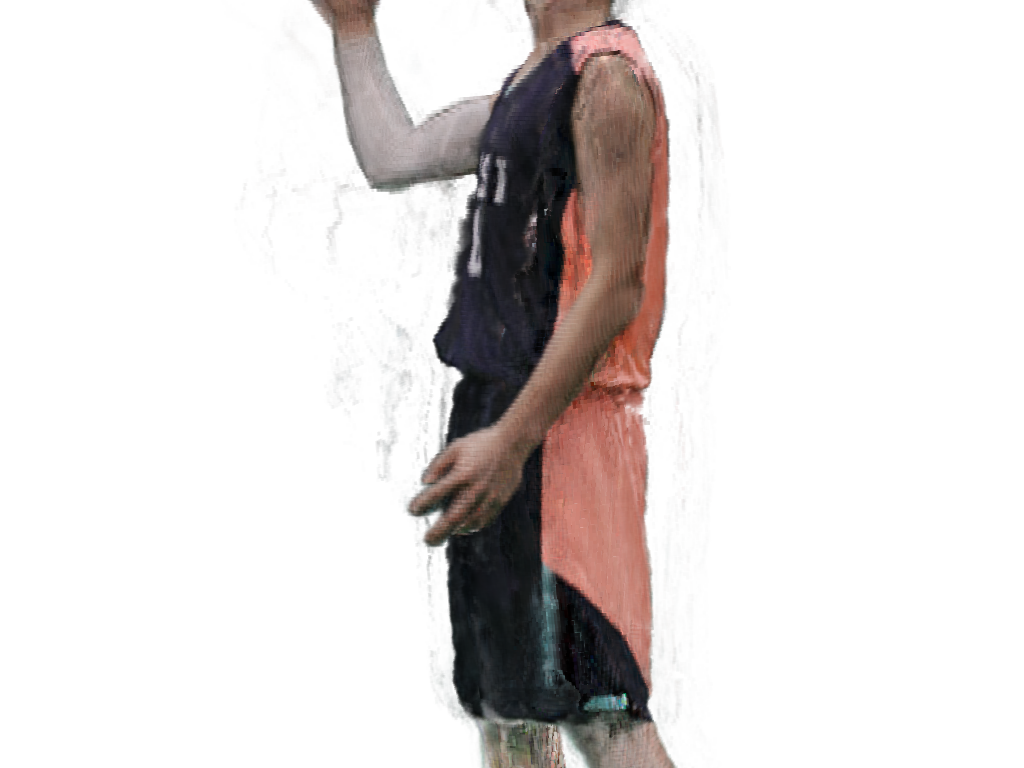}
    \end{minipage}\hfill
    \begin{minipage}{0.18\textwidth}
        \centering
        \includegraphics[width=\textwidth,trim={306 534 462 75},clip]{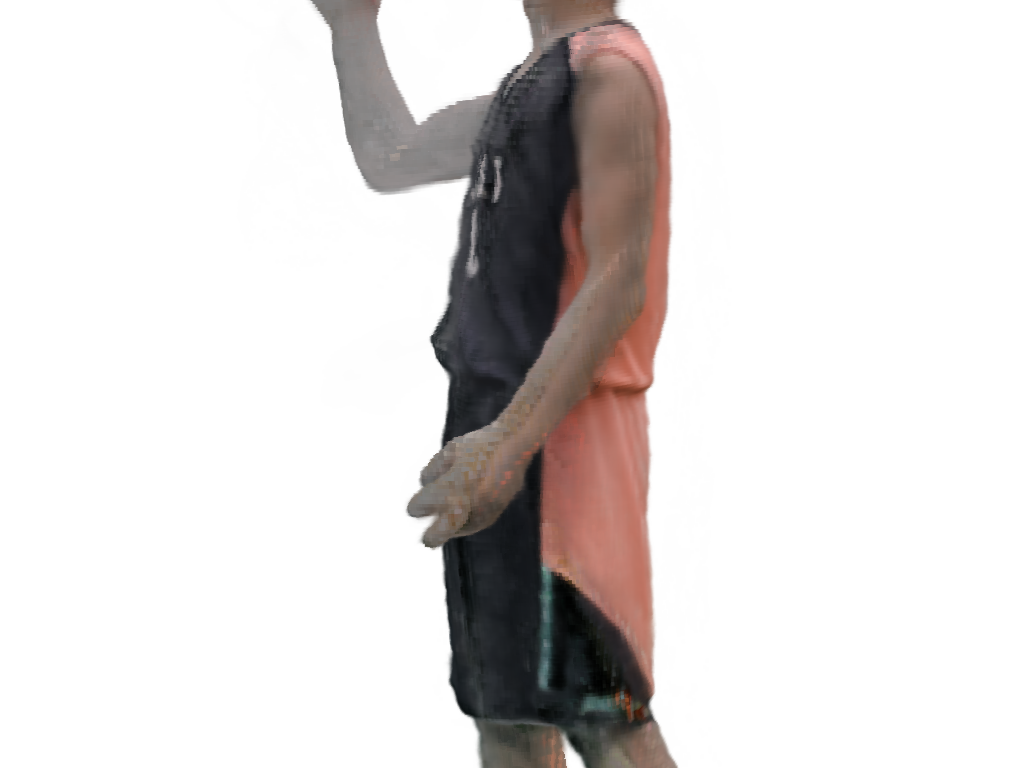}
        \includegraphics[width=\textwidth,trim={306 534 462 75},clip]{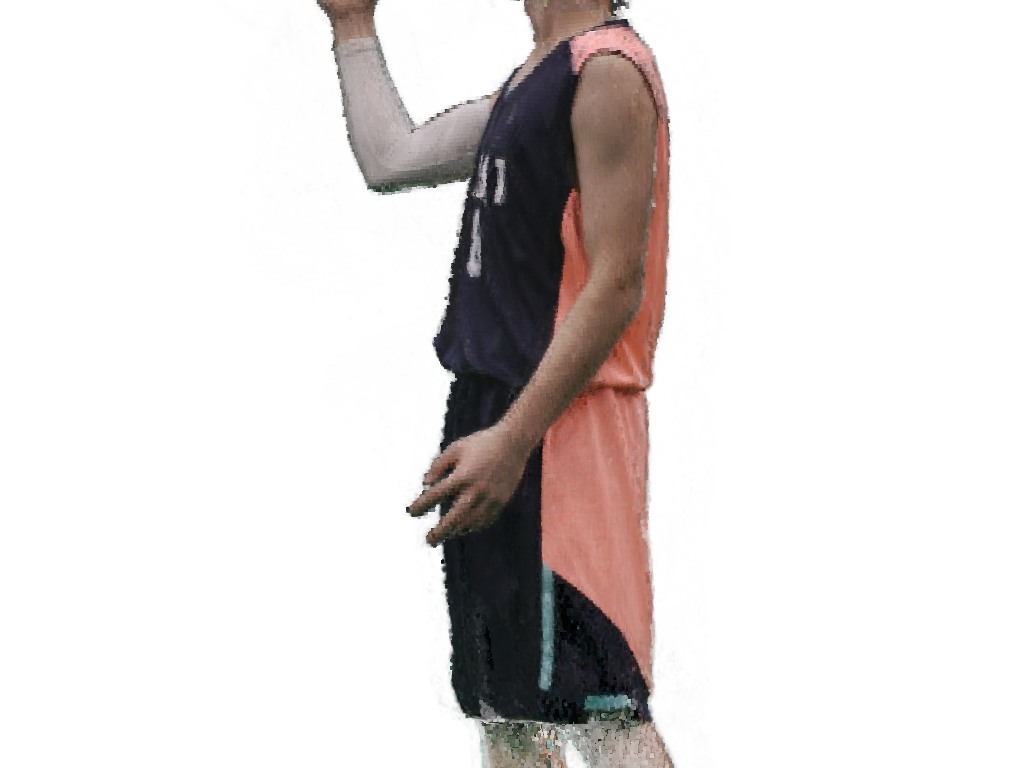}
    \end{minipage}\hfill
    \begin{minipage}{0.18\textwidth}
        \centering
        \includegraphics[width=\textwidth,trim={306 534 462 75},clip]{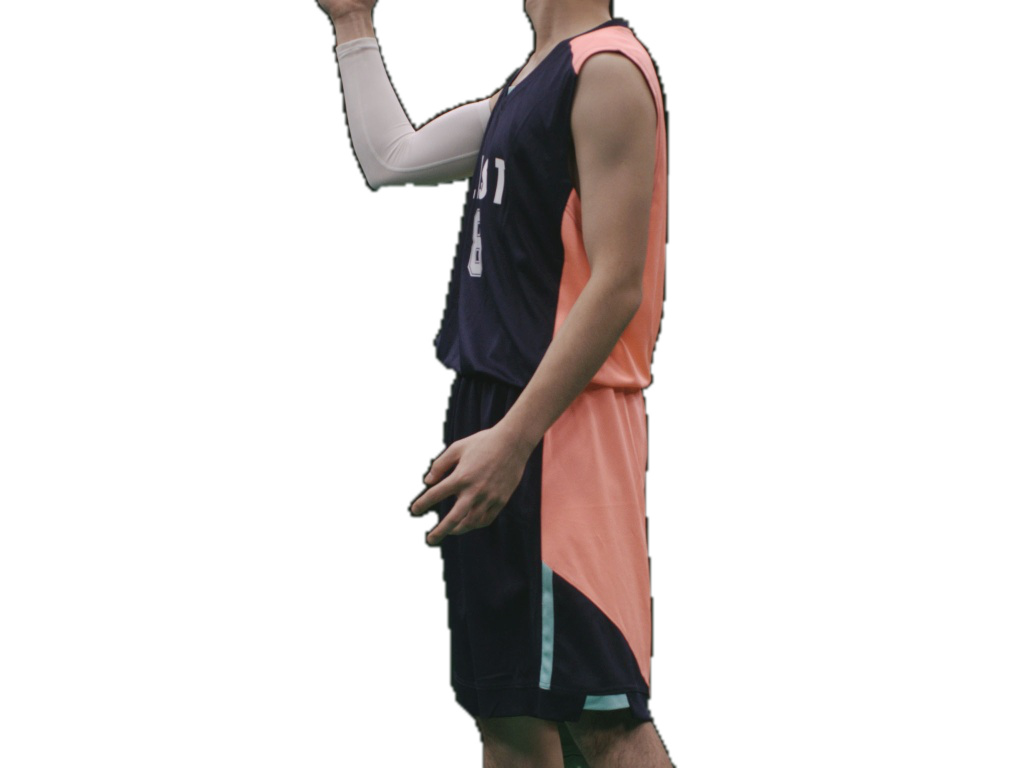}
    \end{minipage}\\
    \begin{minipage}{0.04\textwidth}
        \begin{sideways}Sport 1\end{sideways}
    \end{minipage}\hfill
    \begin{minipage}{0.04\textwidth}
        \begin{sideways}10 epochs \hspace{0.5cm} 0 epochs\end{sideways}
    \end{minipage}\hfill
    \begin{minipage}{0.18\textwidth}
        \centering
        \includegraphics[width=\textwidth,trim={356 371 562 462},clip]{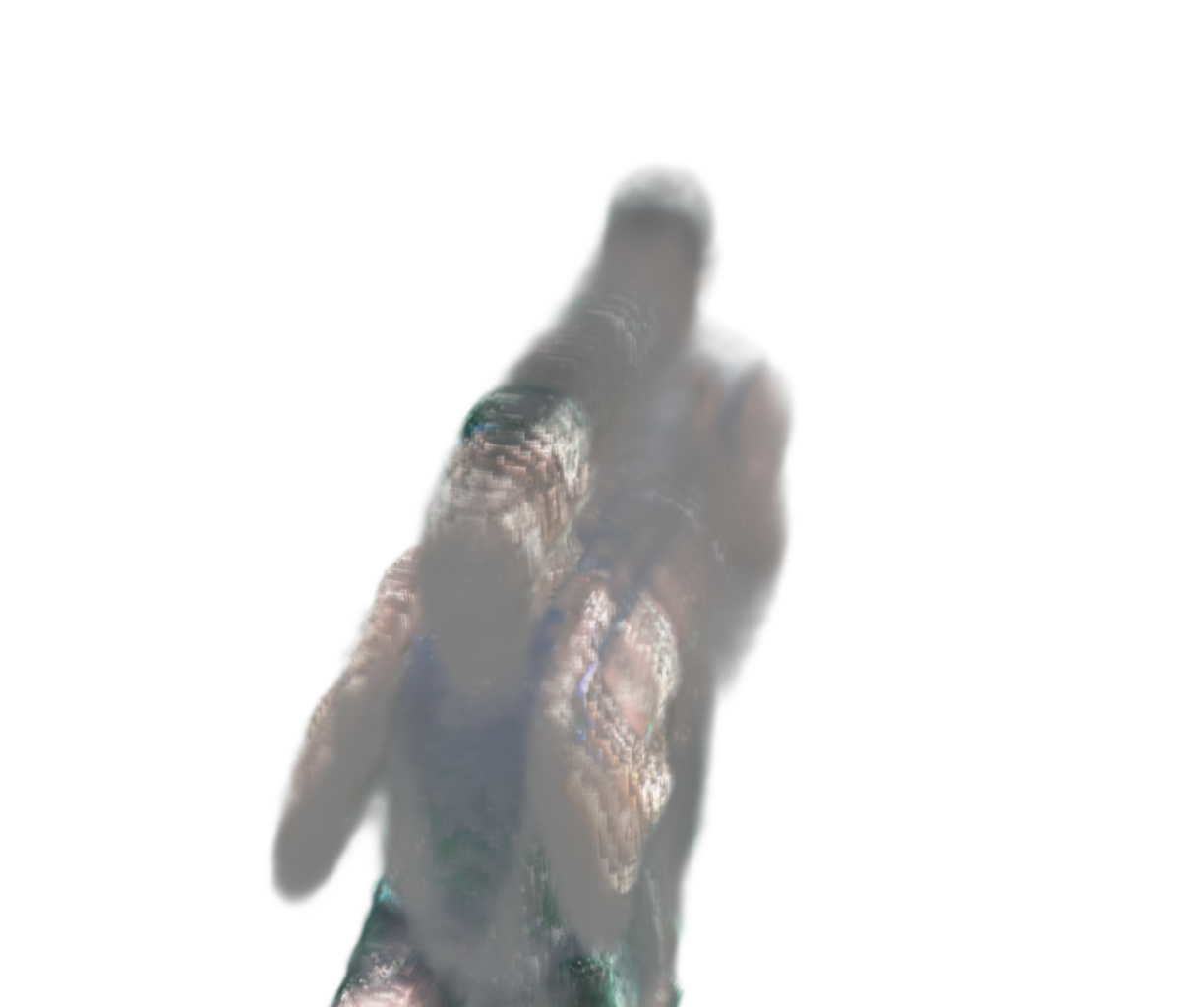}
        \includegraphics[width=\textwidth,trim={356 371 562 462},clip]{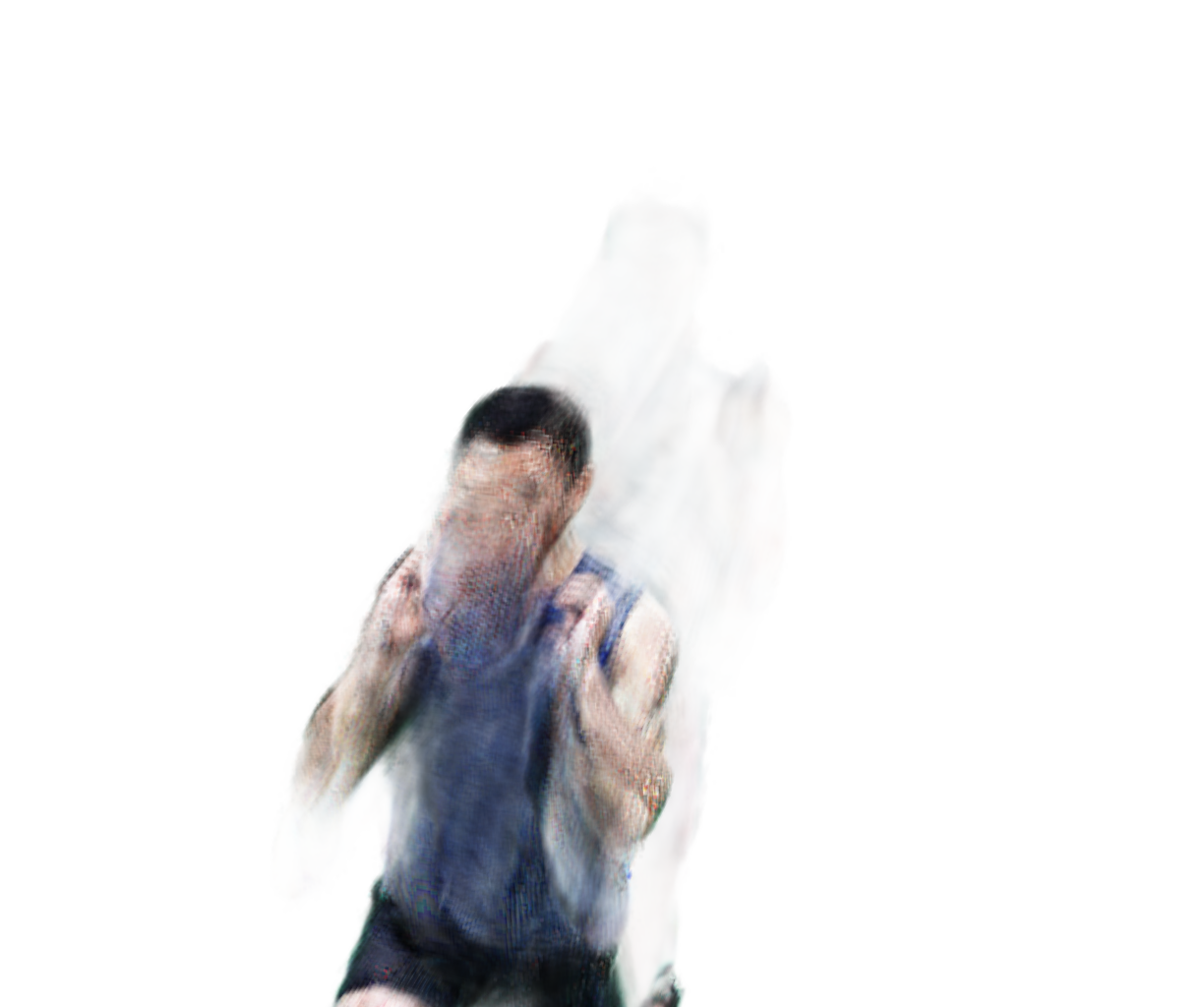}
    \end{minipage}\hfill
    \begin{minipage}{0.18\textwidth}
        \centering
        \includegraphics[width=\textwidth,trim={356 371 562 462},clip]{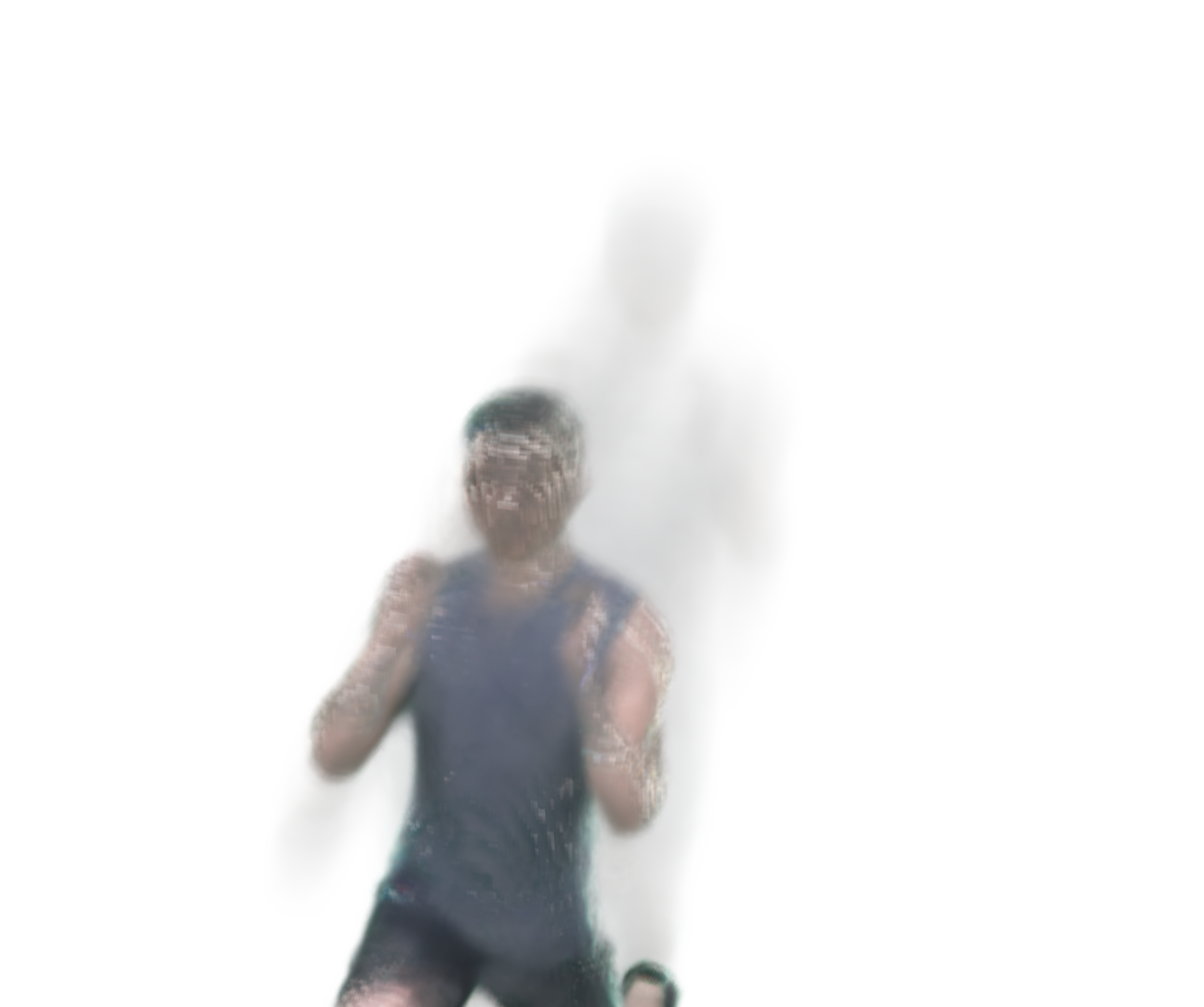}
        \includegraphics[width=\textwidth,trim={356 371 562 462},clip]{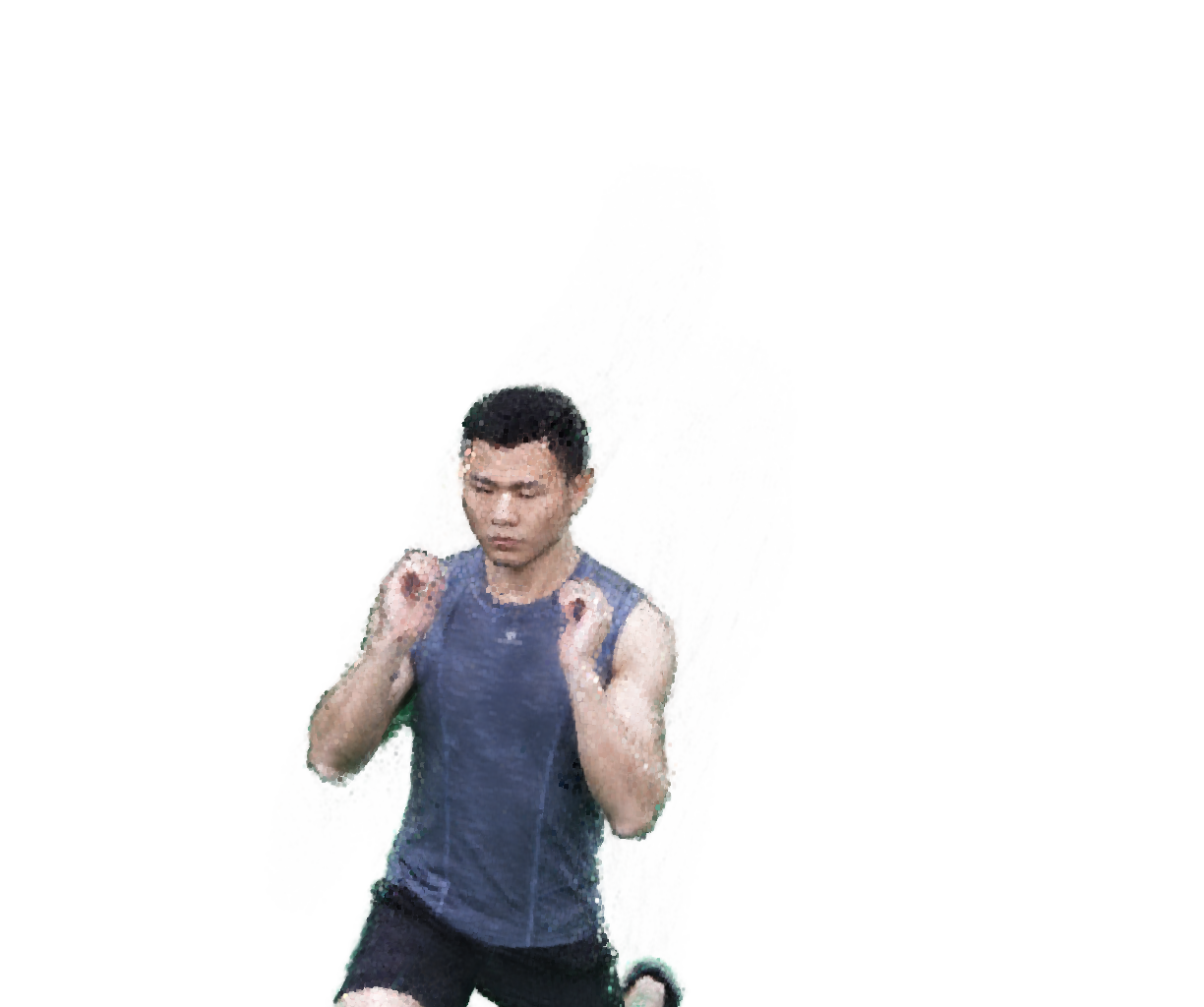}
    \end{minipage}\hfill
    \begin{minipage}{0.18\textwidth}
        \centering
        \includegraphics[width=\textwidth,trim={356 371 562 462},clip]{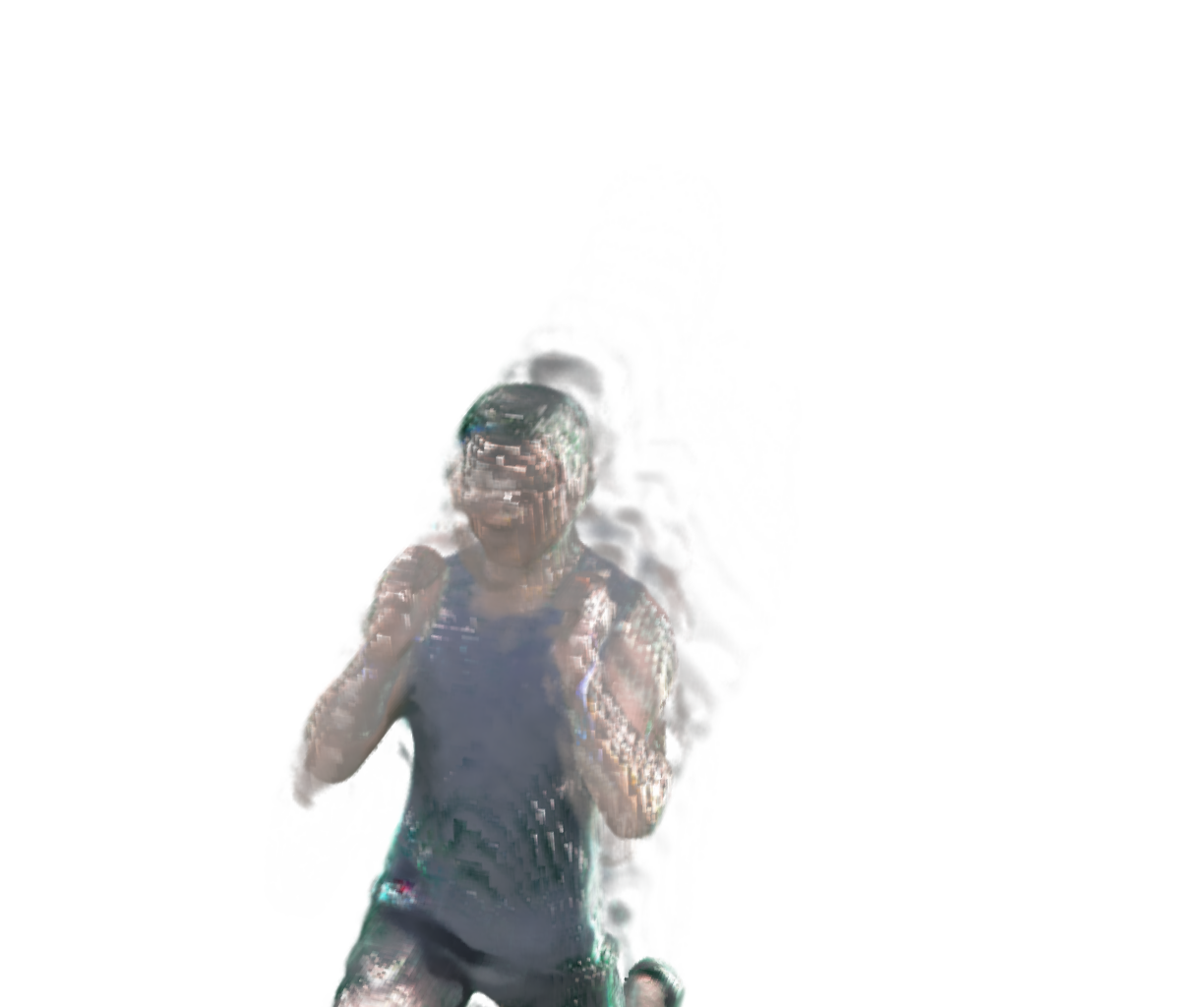}
        \includegraphics[width=\textwidth,trim={356 371 562 462},clip]{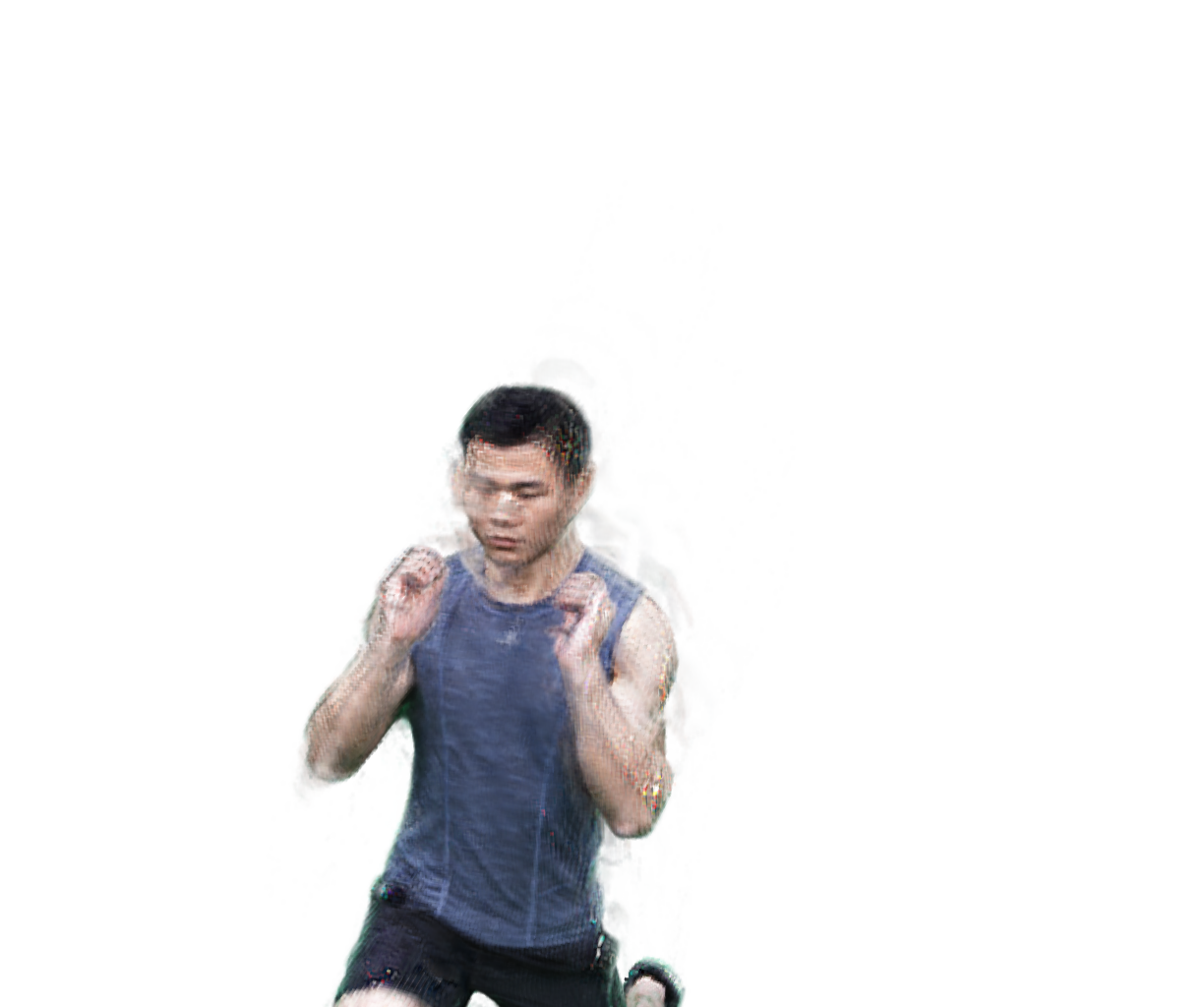}
    \end{minipage}\hfill
    \begin{minipage}{0.18\textwidth}
        \centering
        \includegraphics[width=\textwidth,trim={356 371 562 462},clip]{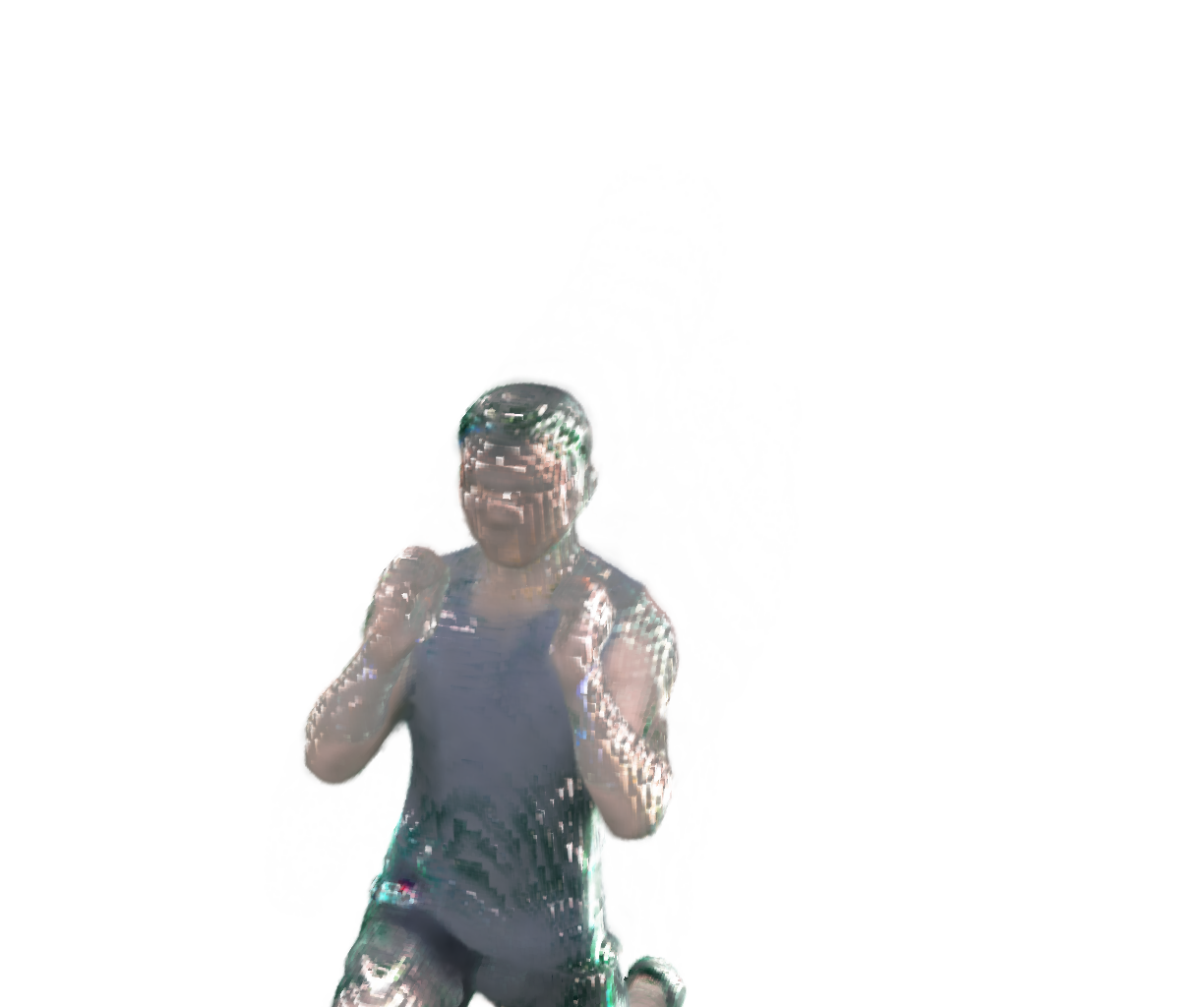}
        \includegraphics[width=\textwidth,trim={356 371 562 462},clip]{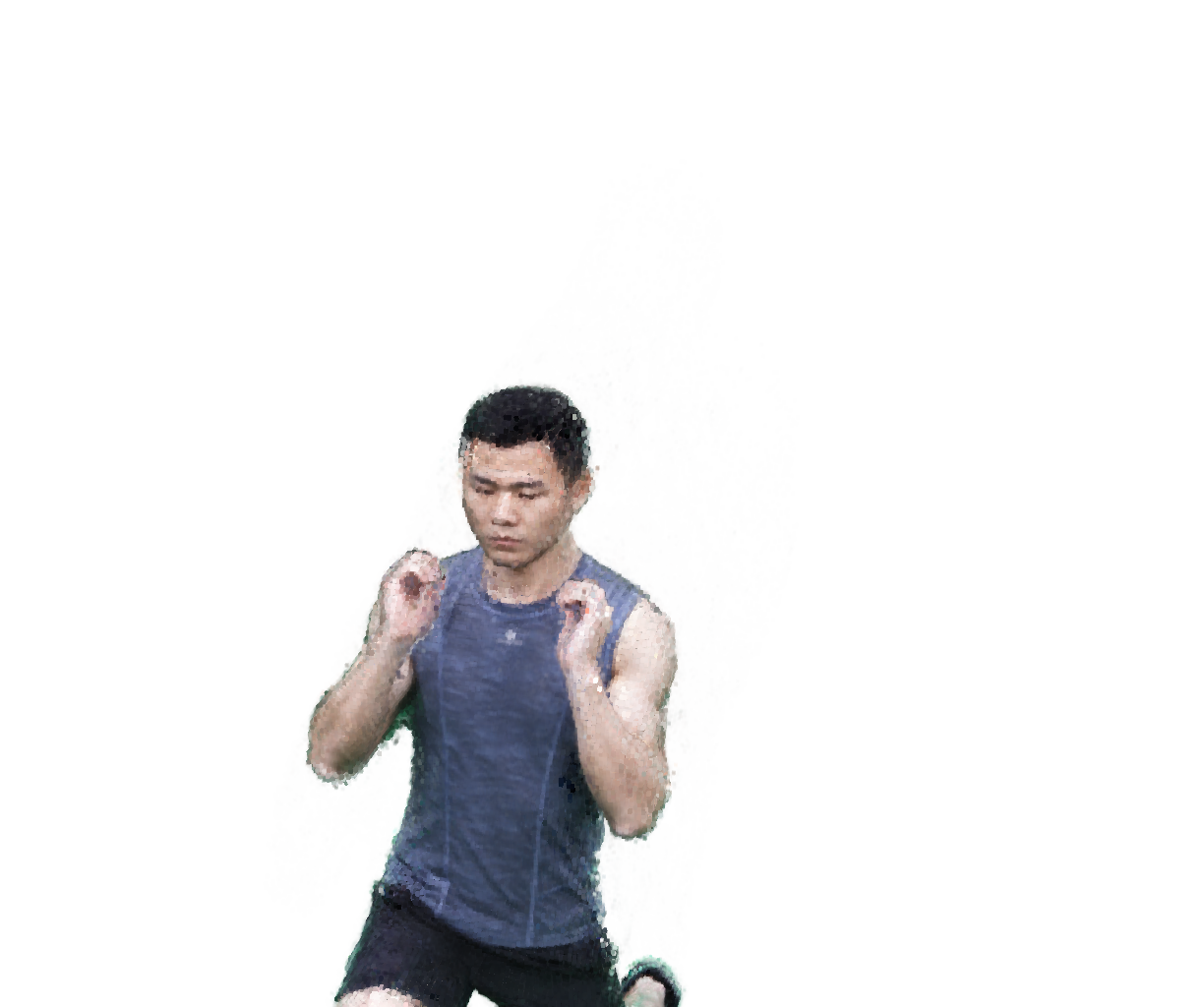}
    \end{minipage}\hfill
    \begin{minipage}{0.18\textwidth}
        \centering
        \includegraphics[width=\textwidth,trim={356 371 562 462},clip]{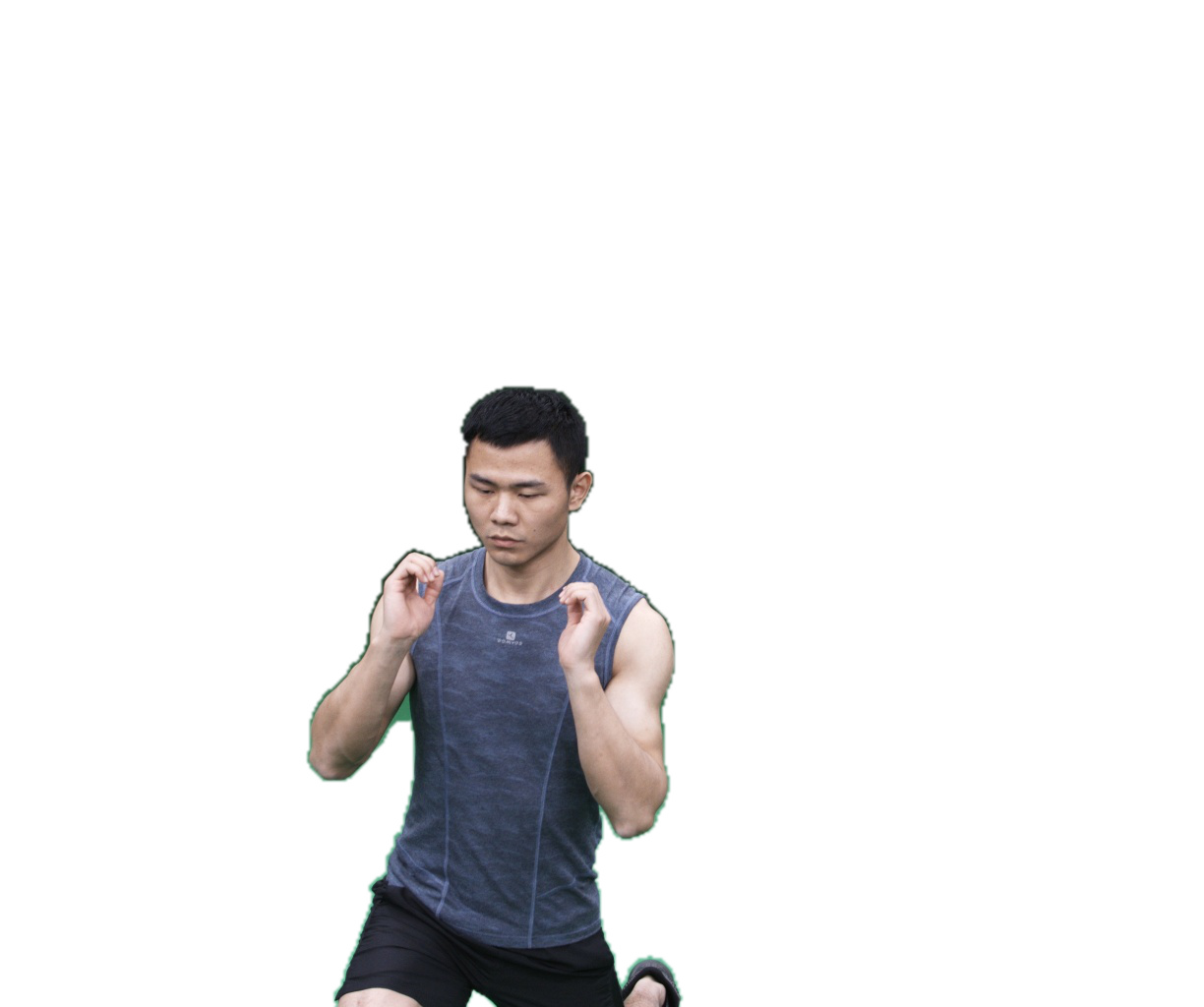}
    \end{minipage}
    \caption{Visual comparison of the ground truth data with the reconstructions of the scenes \textit{Lego}, \textit{Walk}, \textit{Basketball}, and \textit{Sport\,1} with combinations of the logarithmic (log.) and component-dependent (comp.) encoding before and after fine-tuning for 10 epochs.} 
    \label{fig:comparison}
\end{figure*}

Fig.\,\ref{fig:example_renderings} shows renderings of the baseline and our enhanced FPO.
The baseline exhibits artifacts stemming from the geometry of other time steps that are visible as floating structures and are introduced by the Fourier-based compression.
In comparison to FPO-NGP, the amount of these artifacts is significantly reduced with our method.
In the case of fast moving scene content such as the legs in the \textit{Walk} scene, artifacts are mostly removed and much less apparent.
Even without fine-tuning, the geometry is reconstructed well, as can be seen in Fig.~\ref{fig:comparison}.

In the enhanced FPO before fine-tuning gray artifacts are still visible on the reconstruction.
These stem from approximation errors in the SH coefficient functions, which are not altered by our approach.
Leaves that are empty most of the time contain a default value of zero in most static PlenOctrees, which result in the gray coloration.
The fine-tuning process however ensures a realistic reconstruction of RGB colors for all time steps.

\begin{table}[t]
    \centering
    \caption{Comparison of the frame rate [1/s] averaged over all data sets with different combinations of logarithmic encoding (log.) and component-dependent encoding (comp.), both before and after fine-tuning for 1 and 10 epochs. Best and second best results are marked in green and yellow, respectively.}
    \begin{tabular}{lccccccccccccccc}
        \toprule
         Fine-tuning & 0 epochs & 1 epoch & 10 epochs\\
         \midrule
         FPO-NGP& 205.4 & 226.9 & 245.6 \\
         Ours w/o comp.& 157.8 & 222.0 & 266.6 \\
         Ours w/o log.& \colorbox{second}{313.5} & \colorbox{second}{327.7} & \colorbox{second}{339.6} \\
         Ours & \colorbox{best}{349.9} & \colorbox{best}{369.8} & \colorbox{best}{372.6} \\
         \bottomrule
    \end{tabular}
    \label{tab:FPS}
\end{table}

Tab.\,\ref{tab:metrics} provides an overview of the achieved PSNR, SSIM~\cite{wang2004ssim} and LPIPS~\cite{zhang2018lpips} values.
Considering our baseline reimplementation FPO-NGP, we observe a lower performance both qualitatively and quantitatively in comparison to the results of the original reference implementation~\cite{wang2022fourier}.
However, our results are consistent with the evaluations reported in their supplemental material when the generalizable NeRF~\cite{wang2021ibrnet}, which has been specifically fine-tuned on the commercial Twindom data set, is not employed. 
Additional comparisons can be found in the supplementary material.
Besides these observations, our method achieves much better results than the baseline even with only a single epoch of fine-tuning.
Similarly for the case when no further optimization is involved, our method achieves higher metrics than the baseline due to a better initialization of the geometry.
Due to this fact, the creation process of an FPO representation of a dynamic scene is accelerated indirectly, as less time needs to be spent on fine-tuning.

Since our proposed encoding only adds a few additional computation operations, its impact on the performance of the optimization and rendering is minimal.
In fact, we still achieve real-time frame rates, as can be seen in Tab.\,\ref{tab:FPS}.
The resulting FPS are even increased as free space, which previously exhibited artifacts due to the incorrect positive densities, is now correctly identified.
Because free space does not require any computation of color, this step in rendering is now skipped which accelerates rendering significantly.
Furthermore, our encoding does not change the required memory for storing an FPO, which is 2.4\,GiB on average across all tested scenes.

\subsection{Ablation Studies}

In Fig.\,\ref{fig:comparison}, we present a more detailed overview of the effects of each part of our density encoding.
Tab.\,\ref{tab:metrics} lists the corresponding metrics for different scenes.

Especially the logarithmic part improves the reconstruction significantly, since the low frequency approximation can reconstruct the density functions much better than without it and most geometric artifacts are removed or become barely visible.
The DFT and optimization process puts more focus on the reconstruction of lower values and changes between free space and positive densities requiring a smaller error for good results.

The component-dependent part shows to be beneficial for a better initialization of the geometry reconstruction before fine-tuning.
Gray artifacts are removed in most places at most time steps, as zero-densities are represented with negative values and are thus interpreted as free space.
In combination with the logarithmic part, the quality of the initial geometry reconstruction is increased even further.
After fine-tuning, the FPO initialized with only the component-depending part of the encoding achieves better results than the baseline FPO, but also in this case, our full encoding further improves the overall quality.

Both parts of the encoding result in a significant speed-up in rendering before and after fine-tuning, see Tab.\,\ref{tab:FPS}.
While the reconstruction using only the logarithmic part of the encoding shows improved visual quality over the baseline, the rendering speed is decreased without fine-tuning.
Here, zero-densities are better approximated but are still assigned small positive values.
Since higher densities also exhibit smaller values, more values need to be accumulated along the ray to reach the termination criterion which is determined by the transmittance.
Our full encoding consisting of both parts yields the highest frame rate.

We provide additional renderings in the supplemental video and further ablation studies on the number of Fourier coefficients and on the underlying NeRF models in the supplementary material.

\subsection{Limitations}

Similar to other approaches, our method also has some limitations.
The primary focus of our encoding lies in transforming the density functions to make it easier to compress via the Fourier-based signal representation.
SH coefficient functions, however, show different properties and are currently solely compressed using the DFT, which introduces similar artifacts as for the density function.
While fine-tuning allows to improve the reconstruction, obtaining a realistic representation of the colors is also important and a challenging problem.
Furthermore, our methods inherits several limitations of the original FPO approach.
Only the provided data is compressed, so generalization of the scene dynamics in terms of extrapolation and interpolation of the motion is not directly possible.

%% file: sections/conclusion.tex
\section{Conclusion}

In this paper, we revisited Fourier PlenOctrees as an efficient representation for real-time rendering of dynamic Neural Radiance Fields and analyzed the characteristics of its compressed frequency-based representation.
Based on the gained insights of the artifacts that are introduced by the compression in the context of the underlying volume rendering when state-of-the-art NeRF techniques are employed, we derived an efficient density encoding that counteracts these artifacts while retaining the compactness of FPOs and avoiding significant additional complexity or overhead.
Our method showed a superior reconstructed quality as well as a substantial further increase of the real-time rendering performance, and we believe that our insights will also be beneficial for further Fourier-based methods~\cite{wang2023neural}.